\documentclass[sigconf,screen]{acmart}
\usepackage[ruled,vlined]{algorithm2e}
\usepackage{algorithmic}
\usepackage[title]{appendix}
\usepackage{xcolor}
\usepackage{booktabs}
\usepackage{multicol}
\usepackage{mathtools}
\usepackage{multirow}
\usepackage{adjustbox}
\usepackage{pifont}
\usepackage{url}
\usepackage{array}
\definecolor{up_color}{RGB}{0, 108, 0}    
\definecolor{down_color}{RGB}{230, 15, 15}  
\definecolor{darkgreen}{rgb}{0, 0.492, 0}

\newcommand{\changeindicator}[1]{%
  \ifdim #1pt >0pt%
    \textcolor{up_color}{\tiny{\, +#1\%}}%
  \else%
    \textcolor{down_color}{\tiny{\, #1\%}}%
  \fi%
}

\newcommand{\negchangeindicator}[1]{%
  \ifdim #1pt >0pt%
    \textcolor{down_color}{\tiny{\, +#1\%}}%
  \else%
    \textcolor{up_color}{\tiny{\, #1\%}}%
  \fi%
}

\newcommand{\redcross}{\textcolor{red}{\ding{55}}}

\newcommand{\greentick}{\textcolor{darkgreen}{\checkmark}}

\definecolor{darkred}{RGB}{195,16,16}

\AtBeginDocument{%
  }

\setcopyright{acmlicensed}
\copyrightyear{2025} 
\acmYear{2025} 
\setcopyright{cc}
\setcctype{by}
\acmConference[MM '25]{Proceedings of the 33rd ACM International Conference
on Multimedia}{October 27--31, 2025}{Dublin, Ireland}
\acmBooktitle{Proceedings of the 33rd ACM International Conference on
Multimedia (MM '25), October 27--31, 2025, Dublin,
Ireland}\acmDOI{10.1145/3746027.3754961}
\acmISBN{979-8-4007-2035-2/2025/10}

\acmSubmissionID{}

\begin{document}

\title{Multimodal LLMs Can Reason about Aesthetics in Zero-Shot}

\author{Ruixiang Jiang}
\orcid{0000-0001-8666-6767}
\email{rui-x.jiang@connect.polyu.hk}
\affiliation{%
\department{Department of Computing}
  \institution{The Hong Kong Polytechnic University}
  \city{Hong Kong}
  \country{China}
}

\author{Chang Wen Chen}
\orcid{0000-0002-6720-234X}
\email{chen.changwen@polyu.edu.hk}
\affiliation{%
\department{Department of Computing}
  \institution{The Hong Kong Polytechnic University}
  \city{Hong Kong}
  \country{China}
}

\renewcommand{\shortauthors}{Ruixiang Jiang and Chang Wen Chen}

\begin{abstract}
The rapid technical progress of generative art (GenArt) has democratized the creation of \textit{visually appealing} imagery. However, achieving genuine artistic impact – the kind that resonates with viewers on a deeper, more meaningful level – remains formidable as it requires a sophisticated \textit{aesthetic sensibility}. This sensibility involves a multifaceted cognitive process extending beyond mere visual appeal, which is often overlooked by current computational methods. This paper pioneers an approach to capture this complex process by investigating how the reasoning capabilities of Multimodal LLMs (MLLMs) can be effectively elicited to perform aesthetic judgment. Our analysis reveals a critical challenge: MLLMs exhibit a tendency towards hallucinations during aesthetic reasoning, characterized by subjective opinions and unsubstantiated artistic interpretations. We further demonstrate that these hallucinations can be suppressed by employing an evidence-based and objective reasoning process, as substantiated by our proposed baseline, \textbf{ArtCoT}. MLLMs prompted by this principle produce multifaceted, in-depth aesthetic reasoning that aligns significantly better with human judgment. These findings have direct applications in areas such as AI art tutoring and as reward models for image generation. Ultimately, we hope this work paves the way for AI systems that can truly understand, appreciate, and contribute to art that aligns with human aesthetic values. Project homepage: \url{https://github.com/songrise/MLLM4Art}.
\end{abstract}

\ccsdesc[500]{Applied computing~Fine arts}
\ccsdesc[500]{Human-centered computing~Empirical studies in HCI}
\ccsdesc[500]{Computing methodologies~Knowledge representation and reasoning}
\ccsdesc[300]{Computing methodologies~Natural language generation}
\ccsdesc[300]{Computing methodologies~Philosophical/theoretical foundations of artificial intelligence}

\keywords{Computational Aesthetics; Art appreciation and criticism; User modeling; Multimodal LLMs reasoning; Chain-of-thought; Hallucination}


\maketitle

\section{Introduction}
Human aesthetic perception is a multifaceted cognitive process beyond mere visual appeal. It encompasses factors including, but not limited to, uniqueness, narrative, cultural background, and emotional resonance, all of which contribute to the unique aesthetic value of artworks~\cite{gombrich1995story,kant2024critique,Hume1757,joshi2011aesthetics}. The importance of understanding this complex process is widely acknowledged by art practitioners and scholars, as it not only deepens our theoretical understanding of beauty but also informs and inspires art creation~\cite{gombrich1995story,sibley2017aesthetic}. Analogously, in the era of rapidly advancing GenArt, an algorithmic understanding of aesthetics is also crucial to generating truly unique and meaningful artistic expressions~\cite{samo2023artificial,bo2018computational,guo2025can,chamberlain2018putting,hullman2023artificial}.

\begin{figure}
    \centering
    \includegraphics[width=\linewidth]{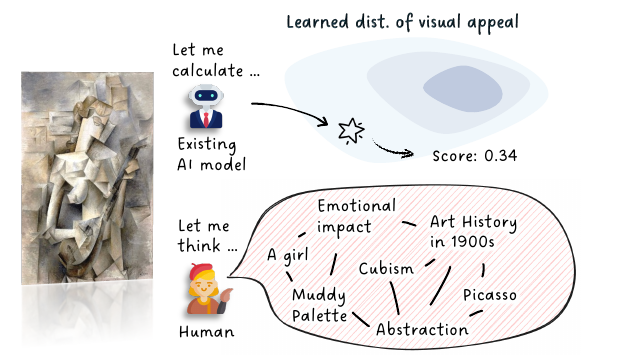}
    \caption{Existing models assess aesthetics as a black-box, which is oversimplified and can misinterpret the aesthetic value. The complex, multifaceted human aesthetic judging process must be captured for human-aligned aesthetic evaluation. Credit: \textit{Girl with a Mandolin (1910), by Pablo Picasso}.}
    \label{fig:teaser}
\end{figure}

Despite its complexity, almost all of the existing computational approaches have oversimplified the concept of aesthetics. These methods predominantly rely on end-to-end learning to obtain black-box scoring models, without engaging with the complex cognitive processes underlying aesthetic judgment. Specifically, systems like traditional Image Quality Assessment (IQA)~\cite{zhu2024adaptive,wang2023exploring}, AI-Generated IQA (AGIQA)~\cite{yang2024moe}, and more recent human preference models (e.g., PickScore~\cite{kirstain2023pick}, VisionPrefer~\cite{wu2024multimodal}) follow this limited paradigm. While these scoring models may effectively assess image quality in some instances, they provide no insights or interpretation for their predictions. More importantly, the measurement of \textit{image quality} or \textit{visual appeal} is a fundamentally superficial understanding of \textit{aesthetics}~\cite{sibley2017aesthetic,kant2024critique,schelling2008philosophy,gombrich1995story,Hume1757,isik2021visual}, and this limited view fosters misalignment with human aesthetic perception in practice~\cite{so2023measuring,ioannou2024evaluation,barnet2015short,chen2024learning,ortlieb2019functional,samo2023artificial,black2023training}. For instance, impactful artworks that are not immediately visually pleasing but involve more in-depth artistic expression may be undervalued by these models. Therefore, modeling the multifaceted cognitive process in human aesthetic judgment is essential to build a truly human-aligned aesthetic evaluator, as illustrated in Fig.~\ref{fig:teaser}.

To capture the underlying process of human aesthetic perception, we operationalize principles from art and philosophy. In particular, the \textit{``formal analysis''}~\cite{barnet2015short,kim2022formal,van2000handbook} in art criticism offers valuable insight that informs our approach. This analysis is a structured method that first describes visual elements in a painting, critically analyzes its formulation, and connects these observations to broader aesthetic principles for objective aesthetic judgment. This analysis extends beyond the subconscious feeling of beauty; it actively weaves perception, emotion, cultural context, and artistic knowledge into a coherent evaluative narrative. In other words, it connects observation and interpretation into words, a holistic capability that bears a striking resemblance to the inference-time reasoning paradigm of recent Multimodal Large Language Models (MLLMs)~\cite{openai2023gpt,team2023gemini,wei2022chain,li2025m2iv,li2025taco}. This parallel presents an opportunity to model the cognitive process behind human aesthetic judgment, inspiring our central research question:

\begin{quote}
\textit{To what extent does reasoning improve the alignment of MLLMs' aesthetic judgments with humans in a zero-shot setting?}
\end{quote}

Answering this question requires understanding both human aesthetic judgment and MLLMs' aesthetic reasoning. We model human judgment through a novel \textbf{tournament-style aesthetic relationship} with philosophy-inspired designs~\cite{tatarkiewicz1963objectivity,kant2024critique,tatarkiewicz2012history} to foster objective comparisons, overcoming the limitation in previous score-based evaluation approaches. Human expert judgment is collected on a newly proposed dataset, \textbf{FineArtBench}, with large-scale semantic labels to facilitate statistically unbiased correlation studies. In parallel, we task MLLMs with the same evaluation. We leverage Chain-of-Thought (CoT) prompting~\cite{wei2022chain} to elicit not just a final decision, but a detailed reasoning trace for why one artwork is preferred over another. Finally, we analyze the alignment between judgments from humans and MLLMs.

Through extensive experiments across different MLLMs, we identify a key challenge of aesthetic reasoning: hallucination. Specifically, we observe a high tendency for MLLMs to hallucinate subjective, unsubstantiated opinions and artistic interpretations. Our analysis further underscores that the bottleneck lies in the reasoning process rather than the base model. To overcome it, we find that evidence-based and objective reasoning is essential to suppress these hallucinations. With these valuable insights, we demonstrate that a simple baseline, \textbf{ArtCoT}, facilitates exceptionally in-depth and multifaceted aesthetic reasoning. This objective reasoning elevates general-purpose MLLMs into state-of-the-art aesthetics evaluators that outperform existing IQA and preference models by a large margin. Furthermore, this powerful aesthetic reasoning has direct applications in human-computer interaction (HCI) and generative art, such as AI art tutors~\cite{ghose2017vincent,xu2024exploration} and reinforcement learning from AI feedback (RLAIF)-enhanced art generation~\cite{black2023training,clark2023directly,guo2025can,wu2024multimodal}, due to its unprecedented interpretability and comprehensiveness.

We summarize our contributions as follows:
\begin{enumerate}
    \item We introduce the task of aesthetic reasoning.

    \item We propose a philosophy-inspired principled approach for objectively modeling human aesthetic judgment.
    \item We introduce \textbf{FineArtBench}, the first large-scale semantically annotated dataset for broad art-related tasks.
    \item Our analysis provides valuable insights into the hallucination bottleneck of MLLMs in aesthetic reasoning.
    \item Our zero-shot baseline \textbf{ArtCoT} points out a promising way for human-aligned aesthetic judgment in AI.
\end{enumerate}

\section{Related Works} 

\subsection{Generative Art, Fine Art and Aesthetics}
At its core, GenArt is a new medium for artistic expression. Neural style transfer (NST)~\cite{deng2022stytr2,huang2017arbitrary,gatys2016image,an2021artflow,jiang2024artist,wang2023nerf,jiang2023avatarcraft} and image generative modeling~\cite{elgammal2017can,rombach2022high,ma2025fine,song2023consistency,lipman2022flow} are two central techniques in this field. Initially, research focused primarily on \textbf{fidelity challenges}, and the goal was to learn the distribution of real artworks, utilizing feature-level metrics such as Gram Loss~\cite{gatys2016image} and FID~\cite{heusel2017gans,wright2022artfid} to quantify the performance. More recently, the focus has shifted towards the \textbf{affective challenge}~\cite{samo2023artificial,li2024playground,wu2024multimodal,hullman2023artificial}, ensuring that the generated results are not only plausible but also visually pleasing. Preference models, such as PickScore~\cite{kirstain2023pick} and HPS~\cite{wu2023human}, are developed to address this challenge. With the rapid scaling of generative models, the acquisition of attractive images has become democratized. This progression allows us to pursue the \textbf{ultimate challenge of art}: striving for artistically meaningful and impactful artworks rather than mere visual appeal. This tri-stage progression was articulated at least 200 years ago by philosophers such as Hegel~\cite{hegel1998aesthetics}. It is contextualized by the transition in art history from technical execution to artistic expression~\cite{gombrich1995story}. Just as aesthetics played a central role in that transition, an algorithmic understanding of aesthetics is key to advancing generative art. This paper directly addresses this issue by proposing a fundamental shift from aesthetic scoring to reasoning to better align with human values.

\begin{figure*}[!t]
    \centering
    \includegraphics[width=\linewidth]{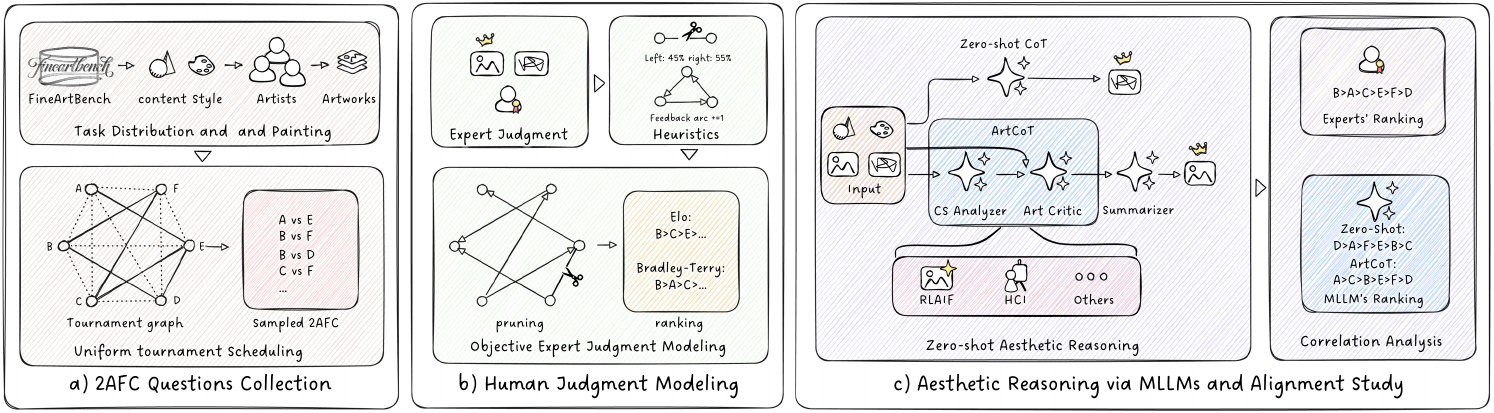}
    \caption{\textbf{Overview of our correlation evaluation pipeline.} (a) First, we generate 2AFC comparison set by sampling content and style from the FineArtBench. (b) Next, human experts perform aesthetic judgment on sampled questions. To ensure annotation consistency, these judgments are filtered with two heuristic indicators and are aggregated as global rankings. (c) In parallel, we leverage different reasoning methods to derive the judgment from MLLMs. Finally, we calculate the correlation coefficient between the MLLM and human rankings as indicators of aesthetic alignment.}
    \label{fig:pipeline}
\end{figure*}

\subsection{Computational Aesthetics} 
Early aesthetic evaluation algorithms are primarily rule-based, assessing the visual quality of paintings or photos using different principles~\cite{bo2018computational,gips1975investigation}. While simple and interpretable, these principles lack the generality to explain all cases. With the emergence of deep learning, data-driven methods~\cite{schuhmann2022laion,hentschel2022clip} attempt to directly learn aesthetic evaluators from image collections labeled with human preferences. In parallel to these explorations, a highly related line of work is developing quantitative metrics for NST. Among them, ArtFID~\cite{wright2022artfid} and Art Score~\cite{chen2024learning} are two notable methods.  Despite the different technical details, all of these black-box models oversimplify the complex aesthetic judgment process and are reportedly misaligned with human perception~\cite{so2023measuring,ioannou2024evaluation,barnet2015short,black2023training}. Recent methods~\cite{zhang2023gpt,huang2024aesexpert,ke2023vila,bin2024gallerygpt,yuan2023artgpt} attempt to incorporate MLLMs for more interpretable and human-aligned aesthetics evaluation. Nevertheless, most require costly instruction-tuning and still focus on visual appeal. This paper focuses on capturing the aesthetics in its full sense by explicitly modeling the aesthetic judgment process.

\subsection{CoT Reasoning}
CoT prompting~\cite{wei2022chain,kojima2022large,hao2024training} is a simple yet powerful technique for reasoning in both LLMs and MLLMs. It functions by requiring the model to decompose the task before arriving at a final answer. It has been proven effective for various tasks that benefit from enhanced logical reasoning, such as math, logic, or commonsense problems. This paper investigates the application of CoT to aesthetics evaluation, a domain that extends beyond strict logical reasoning. Our analysis reveals a special hallucination issue when CoT is applied in aesthetics, and the proposed ArtCoT overcomes this challenge.


\section{Methodology}

\subsection{Problem Definition}
\label{sec:problem}
\textbf{Motivation.} Despite the subjective nature of aesthetic taste, philosophers~\cite{Hume1757,tatarkiewicz2012history,kant2024critique,tatarkiewicz1963objectivity,sircello1968subjectivity} and psychologists~\cite{lindell2011can, palmer2013visual} generally acknowledge a degree of consensus regarding the universality of aesthetic judgments, suggesting the possibility of objective comparative evaluation. This consensus, however, is not absolute. Comparing vastly different artworks, such as \textit{``Mona Lisa''} and \textit{``Guernica''} is much more challenging due to divergent content and style. In contrast, a controlled scenario, such as an art lesson where students draw \textit{porcelain vases} using the \textit{cross-hatching} technique, offers a more robust basis for objective aesthetic comparison. This approach aligns with established practices in art criticism and art education, where rubrics are employed for artwork assessment~\cite{groenendijk2020self,andrade2014formative}. Overall, these observations guide us to explore aesthetic judgment within comparable content and style context for objective comparison.



\textbf{Objective.} To formalize our motivation, we aim to derive a statistically robust global ranking of a set of artists (or art generators) or artworks (or generated art) from human feedback. Formally, let \( \mathcal{C} \), \( \mathcal{S} \), and \( \mathcal{A} \) denote the sets of all content, styles, and artists, respectively (within the studied scope). Each combination \( (c, s) \in \mathcal{C} \times \mathcal{S} \) constitutes a \textit{task} presented to the artists. The resulting artworks from artists \( a_i \in \mathcal{A} \), denoted as \( y_i = a_i(c, s) \), form the set of \textit{candidates} for this task: \( \mathcal{Y}_{c,s} = \{y_1, \ldots, y_k\} \). We then model human aesthetic judgment as a partial ordering on the set \( \mathcal{Y}_{c,s} \):

\begin{equation}    \label{eq:ranks}
    \begin{aligned}
        y_{\pi(1)} \succ  y_{\pi(2)} \succ \cdots \succ y_{\pi(k)},
    \end{aligned}
\end{equation}
where $\pi(i)$ maps the candidate index to its global rank, and $y_1 \succ y_2$ means $y_1$ is judged to be superior than $y_2$. We measure the alignment as the correlation between the ranks constructed from human experts and the MLLM(s), and our target is to study the effectiveness and potential challenges of \textit{reasoning} in enhancing this alignment.

\begin{figure*}
    \centering
    \includegraphics[width=\linewidth]{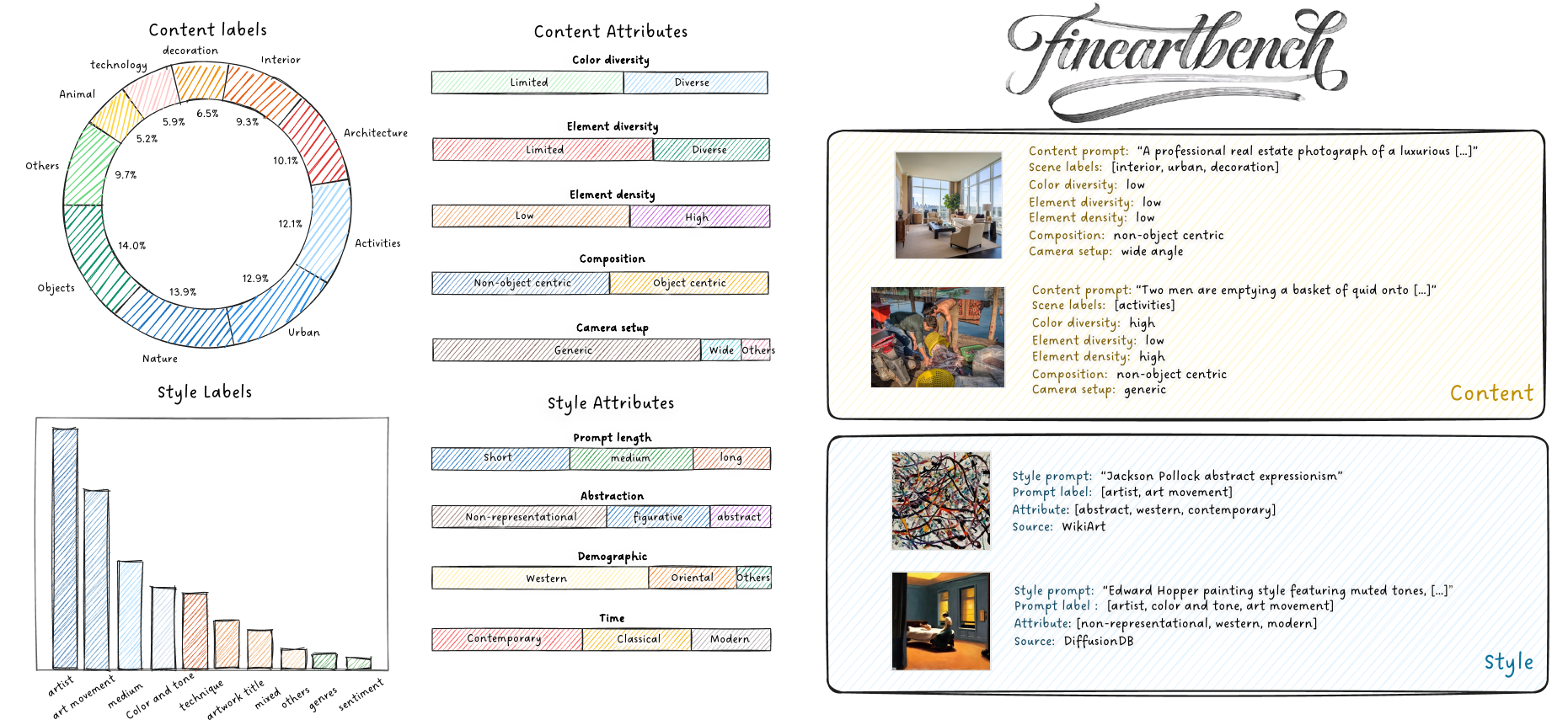}
    \caption{\textbf{The FineArtBench dataset.} Left: The label distribution and semantic attributes for both content and style. Right: examples of two content (top) and two style (bottom). Zoom in for a better view.}
    \label{fig:dataset}
\end{figure*}

\subsection{Modeling Human Aesthetic Judgment}
\label{sec:human_model}
\textbf{Judgment Collection.} Directly obtaining the complete ranking as in Eq.~\ref{eq:ranks} by asking users to sort many items at once is cognitively demanding and can lead to inconsistent judgments. Therefore, inspired by findings in psychometrics~\cite{sullivan2013analyzing,so2023measuring}, we instead employ a two-alternative forced choice (2AFC) task for more efficient judgment collection. Specifically, we iteratively present users with the task specification $(c,s)$, along with two candidate artworks \( (y_i, y_j) \). Users are required to judge the relatively superior artwork without the option to indicate a tie or skip the comparison. We model the judged aesthetic superiority relationship probabilistically:
\begin{equation}
\label{eq:conditional_prefer}
    P(y_i \succ y_j|c,s) = \frac{Q_i}{Q_i + Q_j},
\end{equation}
where \( Q_i \) represents a latent aesthetic competence of \( y_i \), conditioned on the task context of \((c, s)\).

\textbf{Sampling Strategies for 2AFC.} Given the combinatorial complexity of \( \mathcal{C} \times \mathcal{S} \times \mathcal{A} \) ($10^7$ in our case), exhaustive pairwise comparisons are infeasible. Therefore, we generate 2AFC comparisons by sampling, which reduces to a classical tournament scheduling problem~\cite{landau1951dominance}. For a specific candidate set $\mathcal{Y}_{c,s}$, all feasible comparisons can be modeled as complete graph $G(V,E)$, where the node $V \coloneqq \mathcal{Y}_{c,s}$ is the set of candidate results and $E$ stands for pairwise comparisons. We consider two strategies for sampling $E^\prime\subseteq E$, corresponding to two real-world scenarios:
\begin{enumerate}
    \item \textbf{Global sampling:} Uniformly sample an arbitrary number of edges without replacement. This approach is suitable for covering a wider range of content and style to facilitate the ranking of artists across all tasks.

    \item \textbf{Per-task sampling:} Sample edges 
 $|E^\prime|\in[|V|-1,|E|]$ such that the sub-graph is connected with maximum node degree uniformness, meaning that each candidate shall be compared for a similar number of times. This is suitable for determining the ranks of artworks within a specific task.
\end{enumerate}

We design a novel greedy algorithm (described in Appendix~\ref{append: algo}) to efficiently achieve the requirements of per-task sampling.

\textbf{Dealing with Divergent Judgment Annotation.} Due to the inherent subjectivity of aesthetic perception, human judgment can diverge even within the constrained scenario. This variation can arise from inherent annotation noise and is particularly prevalent when the artists perform equally well or poorly. While addressing these challenging edge cases could be valuable, in this paper, given that this is an initial exploration of aesthetic reasoning, we focus on modeling the universality of aesthetics~\cite{kant2024critique,tatarkiewicz1963objectivity,sircello1968subjectivity} with suppressed inter-annotator divergence. This suppression is necessary as it allows us to create an unbiased, reproducible testbed for quantitative evaluation. To this end, we employ the following two heuristics to mitigate annotation subjectivity:

\begin{enumerate}
    \item \textbf{2AFC disagreement:} We exclude the 2AFC questions where judgments tie (i.e., \( P(y_i \succ y_j|c,s) \approx 50\% \)).
    
    \item \textbf{Intransitive relationship:} To deal with intransitive judgment, such as $(a \succ b, b \succ c,c \succ a)$, we apply a feedback arc set (FAS) algorithm~\cite{younger1963minimum} to detect feedback arcs $E^\prime_f$ and drop the whole task
     when $|E^\prime_f|/|E^\prime|\ge \eta$.
\end{enumerate}
    
\textbf{Global Rank Derivation.} We utilize Bradley-Terry (BT) model~\cite{bradley1952rank} and  Elo~\cite{elo1978rating} algorithm to construct Eq.~\ref{eq:ranks} from 2AFC. In the BT model, $Q_i = \operatorname{exp}(\theta_i)$, where the latent parameter $\{\theta_i\}_{i=1}^k$ is optimized with maximum likelihood estimation. The Elo algorithm uses $Q_i = 10^{R_i/400}$, where $R_i$ is the rating of candidate-i that is updated online. The rank of a candidate within a specific task is sorted by its competence $Q_i$, and we treat the average competence among all tasks as the competence of an artist.

\begin{table*}[t]
\centering
\caption{\textbf{Quantitative comparison of aesthetic judgment correlation.} This table presents the statistical correlation (Spearman's $\rho \uparrow$) and significance ($p$ value, $\downarrow$) between rankings generated by different aesthetic evaluators and expert judgments. Both conventional models and MLLMs with different reasoning methods are compared. Performance improvement and decline are calculated as normalized changes relative to the base prompt method for each of the corresponding MLLMs. MLLMs reasoned with ArtCoT consistently demonstrates strong alignment with human judgment.}

\label{tab:model_comparison}
\resizebox{\linewidth}{!}{
\begin{tabular}{ll|cccccccc}
\toprule
\multirow{3}{*}{Model} & \multirow{3}{*}{Reasoning} & \multicolumn{4}{c|}{Per-Artist Alignment} & \multicolumn{4}{c}{Per-Task Alignment} \\
\cmidrule{3-10}
& & \multicolumn{2}{c}{Elo} & \multicolumn{2}{c}{Bradley-Terry} & \multicolumn{2}{c}{Elo} & \multicolumn{2}{c}{Bradley-Terry} \\
& & $\rho$ $\uparrow$& $p$-value $\downarrow$& $\rho$ $\uparrow$& $p$-value $\downarrow$& $\rho$ $\uparrow$& $p$-value $\downarrow$& $\rho$ $\uparrow$& $p$-value $\downarrow$\\
\midrule
Random guess & -- & 0.061 & 0.896 & 0.059 & 0.855 & 0.068&0.153 &0.026 & 0.290 \\

CLIP-IQA\cite{wang2023exploring} & -- & 0.366& 0.038 & 0.321 & 0.091 & 0.324&0.004 & 0.314 &0.002\\
Compare2Score~\cite{zhu2024adaptive} & -- & 0.091 &  0.940 & 0.085  & 0.950 & 0.178 & 0.042& 0.173 & 0.119  \\
CLIP Score~\cite{radford2021learning} & -- & -0.024 & 0.945 & 0.030 & 0.951 &-0.333 &0.347 &0.006 & 0.987  \\
PickScore~\cite{kirstain2023pick} & -- & 0.275& 0.192& 0.329&0.063 & 0.193 & $<10^{-3}$ & 0.201 & 0.001 \\
HPSv2~\cite{wu2023human} & -- & 0.343&0.074 &0.352 & 0.056& 0.321 & $<10^{-3}$ & 0.285 & 0.007\\

Aesthetics Predictor~\cite{schuhmann2022laion} & -- & 0.424 & 0.016 & 0.456 & 0.008 & 0.385 &$<10^{-3}$ & 0.404 & $<10^{-3}$\\
\midrule

GPT-4o & Base & 0.395 & 0.084 & 0.432 & 0.023 & 0.328&0.003 &0.331 &0.006 \\
Claude 3.5-sonnet & Base & 0.341 & 0.032&0.217 &0.236 & 0.312 & $<10^{-3}$  & 0.367 & $<10^{-3}$ \\
Gemini 1.5-flash & Base & 0.496& 0.003 & 0.513 &0.001 & 0.479 & $<10^{-3}$&0.353 & $<10^{-3}$\\

\midrule
GPT-4o & Zero-shot CoT & 0.415\changeindicator{3}& 0.034&0.487\changeindicator{10} &0.015 & 0.299\changeindicator{-4}&0.097 & 0.313\changeindicator{-3}& 0.031\\
Claude 3.5-sonnet & Zero-shot CoT & 0.293\changeindicator{-7} & 0.177 & 0.264\changeindicator{6}& 0.174 & 0.108\changeindicator{-30} & 0.068 & 0.081\changeindicator{-45} &0.082 \\
Gemini 1.5-flash & Zero-shot CoT & 0.206\changeindicator{-58} & 0.292 & 0.272\changeindicator{-50}& 0.142 &0.376\changeindicator{-20}& $<10^{-3}$ & 0.327\changeindicator{-4} & $<10^{-3}$ \\

\midrule
GPT-4o & ArtCoT & 0.630\changeindicator{39} &0.003 &0.721\changeindicator{51}&0.001 & 0.591\changeindicator{39}&  $<10^{-3}$ & 0.548\changeindicator{32}& $<10^{-3}$ \\

Claude 3.5-sonnet & ArtCoT & 0.598\changeindicator{39} & 0.009 & 0.546\changeindicator{42} & 0.016& 0.492\changeindicator{26} &$<10^{-3}$ &0.487\changeindicator{19} &$<10^{-3}$ \\

Gemini 1.5-flash & ArtCoT & \textbf{0.705}\changeindicator{41} & $<10^{-3}$ & \textbf{0.741}\changeindicator{47}& $<10^{-3}$ & \textbf{0.624}\changeindicator{28}& $<10^{-3}$&\textbf{0.577}\changeindicator{35} & $<10^{-3}$\\

\midrule

Claude-3.7-sonnet-thinking& (Built-in) & 0.285 & 0.225& 0.154 &0.481 & 0.171 & 0.014  & 0.175 & 0.009 \\
Gemini-2.0-flash-thinking& (Built-in) & 0.204 & 0.262&0.254 &0.191 & 0.157 & 0.004  & 0.245 & 0.003 \\
\bottomrule
\end{tabular}}
\label{tab:main_result}
\end{table*}

\subsection{Zero-shot Aesthetics Reasoning via ArtCoT}
\label{sec:art_cot}

Recent inference-time scaling paradigms, particularly the CoT prompting~\cite{wei2022chain,kojima2022large} demonstrate impressive zero-shot task-solving capability. However, CoT is more widely studied for commonsense and logical reasoning. Applying it to aesthetic judgment, which is not a direct logical task, remains an unexplored direction. Very recently, research suggest the alignment challenge in this application may differ from logical tasks~\cite{sprague2024cot,liu2024mind}. In practice, the key issue we identified is hallucination, where judgments are made prematurely, relying on subjective and superficial justifications. In other words, during aesthetic reasoning, \textbf{MLLMs inherently tend to ``feel'' rather than to ``reason''}, and this \textbf{hallucinated feeling} is tied with severe degradation of the MLLM's aesthetic judgment with human alignment. We detail this issue in the Sec.~\ref{sec:discuss_hallucination}.

Motivated by the formal analysis in art criticism ~\cite{barnet2015short,kim2022formal,van2000handbook}, our key insight is to explicitly ground the MLLMs' decision-making with evidence and domain knowledge in art, so as to reduce spurious reliance on hallucinated feelings. Based on this principle, we propose a simple yet effective baseline, \textbf{ArtCoT}, which frames the multifaceted aesthetic judgment as an evidence-based reasoning process.

ArtCoT features a two-stage reasoning process. In the initial stage, the MLLM acts as a \textbf{CS Analyzer}, generating a detailed and concrete analysis of the input task $(c,s)$, as well as the paired 2AFC images $(y_i,y_j)$. This stage provides the context for aesthetic judgment.  In the second phase of \textbf{Art Critic}, the MLLM is prompted to critically evaluate its previous observation and to \textit{synthetically judge}~\cite{kant1950prolegomena,kant2024critique} the aesthetic value. The analysis in this stage is multifaceted, with discussion not limited to stylistic features of $s$, aesthetic principles, technical execution of the two candidates, and potential emotional impact. Overall, these two stages link context and world knowledge into the aesthetic judgment, fundamentally different from the existing paradigm based on vision features alone. Effectively, it establishes a rubric~\cite{groenendijk2020self,andrade2014formative}, translating the subjective perception of beauty into more well-defined sub-tasks that can be logically reasoned~\cite{christensen2016dimensions,coleman1966phenomenology,li2023theme}.

Importantly, we aim to capture this reasoning trace as a model of the cognitive process behind aesthetic judgment, instead of simply arriving at the 2AFC result. This trace is a rich representation and is highly interpretable, which can benefit a wide range of downstream applications. As this paper focuses on assessing the alignment between MLLMs and humans, we further employ a summarizer that takes the reasoning trace as input and outputs a binary answer for the winner. The overall workflow is visualized in Fig.~\ref{fig:pipeline}.

\subsection{FineArtBench}
\label{sec: fineartBench}
The collection of content and style (i.e., $\mathcal{C}, \mathcal{S}$) is most widely studied in NST literature~\cite{jiang2024artist,huang2024diffstyler,deng2022stytr2}. However, the majority of these papers rely on ad-hoc sampling from existing datasets, such as MS-COCO~\cite{lin2014microsoft} and WikiART~\cite{wikiart}. This ad-hoc approach has two critical limitations: (1) \textbf{Limited Scope and Inherent Bias}, usually covering a small range of art styles that may lead to bias in aesthetics evaluation. (2) \textbf{Lack of Semantic Annotation}: the lack of semantic-level labeling of style and content impedes a fine-grained understanding of the methods' strengths and weaknesses.

To circumvent these challenges, we propose \textbf{FineArtBench}, a dataset for GenArt with unprecedented \textbf{scale, diversity and quality}. In terms of scale, it contains 1,000, and 1,000 densely annotated content and styles, respectively. The data is harvested from diverse sources~\cite{kirillov2023segment,lin2014microsoft,wikiart,wang2022diffusiondb}, and we use MLLMs, enabling comprehensive comparison for downstream art-related tasks. Moreover, we provide content and style in two modalities (text and image), further expanding its versatility. Second, we perform a semi-automated annotation process (detailed in Appendix~\ref{supp:dataset_detail}) to provide semantic information, such as scene in content (e.g., \textit{nature, portrait}) and abstractness of style (e.g., \textit{figurative, abstract}).  These annotations enable an unprecedented, fine-grained analysis of model performance. A comparison of FineArtBench with existing datasets is presented in Tab.~\ref{tab:dataset_comparison}, and Fig.~\ref{fig:dataset} presents an overview of FineArtBench.

\begin{table}[]
\centering
        \caption{\textbf{Comparison of representative GenArt benchmark datasets.} The proposed FineArtBench offers significantly more content and style instances, with fine-grained multimodal annotations.}
        
     \resizebox{\linewidth}{!}{
    \begin{tabular}{c|cccc}
    \toprule
        Dataset/Protocol & \# Content &\# Style&Multimodal&Semantic Labeling \\
         \midrule           
         StyleID~\cite{chung2024style} & 20 &40 &\redcross&\redcross \\
         LAION-Aesthetics~\cite{schuhmann2022laion} & -&$\sim$50&\redcross&\redcross\\
        ArtBench~\cite{liao2022artbench}&-&10&\redcross&\redcross\\

         StyleBench~\cite{gao2024styleshot}&20&73&\greentick&\redcross\\
        AGIQA-3k~\cite{gao2024styleshot}& $\sim300$ & 5&\greentick&\redcross\\
         \midrule
         \textbf{FineArtBench}&\textbf{1000}&\textbf{1000}&\greentick&\greentick\\\bottomrule
    \end{tabular}}

    \label{tab:dataset_comparison}
\end{table}

\section{Experiments}
\subsection{Experiment Setup}

\textbf{Stylization Models.} Given the size of $|\mathcal{C}\times \mathcal{S}|$, hiring real human artists for scalable dataset collection is unfortunately infeasible: a single painting requires days of expertise. On the other hand, existing publicly available artworks have a limited style that only covers a small subset of $\mathcal{S}$. Even if we collect some paired paintings, the MLLMs are prone to memorization rather than reasoning~\cite{bin2024gallerygpt}. To this end, we primarily rely on neural style transfer (NST) models as our ``artists.''  We evaluated a total of 10 NST models, all executed with the default configurations suggested by respective authors. The methods include AdaIN~\cite{huang2017arbitrary}, ArtFlow~\cite{an2021artflow}, ControlNet~\cite{zhang2023adding}, DDIM~\cite{song2020denoising}, DiffArtist~\cite{jiang2024artist}, DiffStyler~\cite{huang2024diffstyler}, InstantStyle~\cite{wang2024instantstyle}, Instruct-pix2pix~\cite{brooks2023instructpix2pix}, StyleID~\cite{chung2024style}, and Sty-Tr2~\cite{deng2022stytr2}. These stylization models encompass a wide range of architectures and exhibit varying degrees of stylization capability. This diversity provides a robust testbed for evaluating the performance of MLLMs in art evaluation.

\textbf{Alignment Metrics.} Following~\cite{jiang2024artist,chen2024learning,zhang2023gpt}, we use Spearman's correlation coefficient~\cite{spearman1987proof} to quantify the alignment between judgment from experts and MLLMs. A Spearman's $\rho$ closer to 1 indicates a stronger positive linear correlation of ranking, which suggests better alignment. To ensure robustness, we calculate the average $\rho$ from five random and independent splits for the global (per-artist) sampling scheme, while for per-task sampling, we correlate each independent task. Considering the size of the independent test and typical $p$ values, we combine $p$ using Pearson's and Fisher's methods for per-artist and per-task scenarios, respectively, as suggested by~\cite{heard2018choosing}. A lower $p$ value means stronger statistical significance.

\textbf{Compared Methods.} As the first exploration on MLLMs for aesthetic reasoning, finding direct comparisons can be challenging. Therefore, we broadly compare three groups of methods. (a), \textbf{IQA methods:} including CLIP-IQA~\cite{wang2023exploring} and Compare2Score~\cite{zhu2024adaptive}; (b) \textbf{Text-image alignment and preference scores:} CLIP Score~\cite{radford2021learning}, Aesthetic Predictor~\cite{schuhmann2022laion}, PickScore~\cite{kirstain2023pick} and HPSv2~\cite{wu2023human}.  Despite ArtScore~\cite{chen2024learning} sharing the most similar objective with this paper, we exclude it due to the unavailability of code at the time this paper was written. We also consider (3) \textbf{Zero-shot MLLMs}, where three main-stream MLLMs are inspected, including GPT-4o~\cite{openai2023gpt}, Gemini 1.5~\cite{team2023gemini}, and Claude 3.5~\cite{claude}. We apply different reasoning methods on top of them, including \textit{base} (directly output the winner), \textit{zero-shot CoT}~\cite{wei2022chain}, and the proposed \textit{ArtCoT}. We also consider recent reasoning MLLMs, including \verb|Claude-3.7-sonnet-reasoning| and \verb|Gemini-2.0-flash-thinking|. The experiment of all MLLMs is repeated thrice with default hyperparameters to ensure robustness.

\subsection{Human Judgment Collection}
We recruit 18 human experts with general knowledge of fine art for annotation. We collected a total of 80,000 initial judgments. For per-task sampling, we sample $\mathcal{O}(klog(k))$ for each task. We prune uncertain human feedback with $P(y_i\succ y_j|c,s)\in [0.4, 0.6]$, and remove intransitive task with $\eta>0.15$. As a result, 20.1\% of the feedback is filtered out due to pairwise divergence, while the remaining 15.3\% is removed due to a high intransitivity. To quantify the inter-annotation agreement, we use McFadden's $R^2$~\cite{mcfadden1972conditional} to assess the transitivity of global ranking constructed from BT and Elo, reported in Tab.~\ref{tab:agreement}. An $R^2 \ge0.2$ is generally considered a good fit, which implies the judgments are statistically consistent. Further details are in Appendix~\ref{supp:human_detail}.

\begin{table}[]
    \caption{Inter-Annotator Agreement. Higher goodness-of-fit $R^2$ indicate more concordant global ranking. }
    \label{tab:agreement}
    \centering
    \begin{tabular}{ccccc}
    \toprule
         & Raw  & + prune uncertain  & + drop non-transitive\\\midrule
 
         $R^2 \uparrow$& 0.17& 0.26 & 0.31 \\
    \bottomrule
    \end{tabular}
\end{table}

\subsection{Main Result: Human Alignment}

We compare the correlation and statistical significance of all compared methods under two sampling strategies. The results in Tab.~\ref{tab:main_result} demonstrate that aesthetic reasoning substantially enhances an MLLM's aesthetic judgment, achieving average improvements of 44\%, 30\% in the per-artist and per-task setup, respectively. Compared with other specialized IQA or preference models, MLLMs prompted with ArtCoT demonstrate state-of-the-art alignment even without task-specific optimization (i.e., in zero-shot). Interestingly, the zero-shot CoT adversely affects alignment, decreasing it by 16\% and 18\% on average. Overall, these results indicate that (1) reasoning can indeed make MLLMs human-aligned aesthetic evaluators, conditioned on (2) proper elicitation of their reasoning capability. Next, we analyze these findings in detail.

\section{Analysis and Discussion}\label{sec:analysis}

\begin{table*}[!t]
    \centering
    \caption{\textbf{Response subjectivity and hallucination from different prompting methods.} We report the average length of thinking trace (in words), two subjectivity metrics, and human-annotated hallucination score (normalized) from four perspectives. Reasoning traces generated with ArtCoT contain less subjectivity and hallucination. (*): Trace from Claude-3.7-sonnet-thinking. }
    \resizebox{\linewidth}{!}{
    \begin{tabular}{cccc|cc|cc}
    \toprule
    \multirow{2}{*}{Method} & \multirow{2}{*}{Avg. length} &\multicolumn{2}{c|}{\textbf{Response Subjectivity}} & \multicolumn{2}{c|}{\textbf{Hallucination (Factuality)}} & \multicolumn{2}{c}{\textbf{Hallucination (Extrinsic)}} \\
     \cmidrule(lr){3-4} \cmidrule(lr){5-6} \cmidrule(lr){7-8}
    &  & TextBlob $\downarrow$ & Word frequency ($\%$)  $\downarrow$ & Visual $\downarrow$ & Textual $\downarrow$& Opinionated $\downarrow$ & Interpretation $\downarrow$\\
    \midrule
    Thinking MLLM (*) & 578.71 & 0.44 & 11.41 & 0.21& 0.15 & 0.44 & 0.49 \\
    GPT, 0-shot & 71.73 & 0.44 & 17.34 & 0.31 & 0.17 & 0.45 & 0.33 \\
    Claude, 0-shot & 223.51 & 0.44 & 20.15 & 0.27 & 0.22 & 0.58 & 0.48 \\
    Gemini, 0-shot & 123.07 & 0.46 & 21.05 & 0.31 & 0.17 & 0.54 & 0.59 \\
    \midrule
    GPT, ArtCoT & 384.45 & 0.30\negchangeindicator{-25} & 5.04\negchangeindicator{-14} & 0.29\negchangeindicator{-3} & 0.19\negchangeindicator{2} & 0.32\negchangeindicator{-24} & 0.18\negchangeindicator{-22} \\
    Claude, ArtCoT & 580.21 & 0.29\negchangeindicator{-26} & 6.15\negchangeindicator{-18} & 0.31\negchangeindicator{5}& 0.13\negchangeindicator{-12} & 0.22\negchangeindicator{-85} & 0.31\negchangeindicator{-33} \\
    Gemini, ArtCoT & 740.21 & 0.23\negchangeindicator{-43}& 6.51\negchangeindicator{-18} & 0.28\negchangeindicator{-4} & 0.19\negchangeindicator{5} & 0.17\negchangeindicator{-80} & 0.22\negchangeindicator{-53} \\
    \bottomrule
    \end{tabular}}
    \label{tab:subjective}
\end{table*}

\begin{figure}
    \centering
    \includegraphics[width=\linewidth]{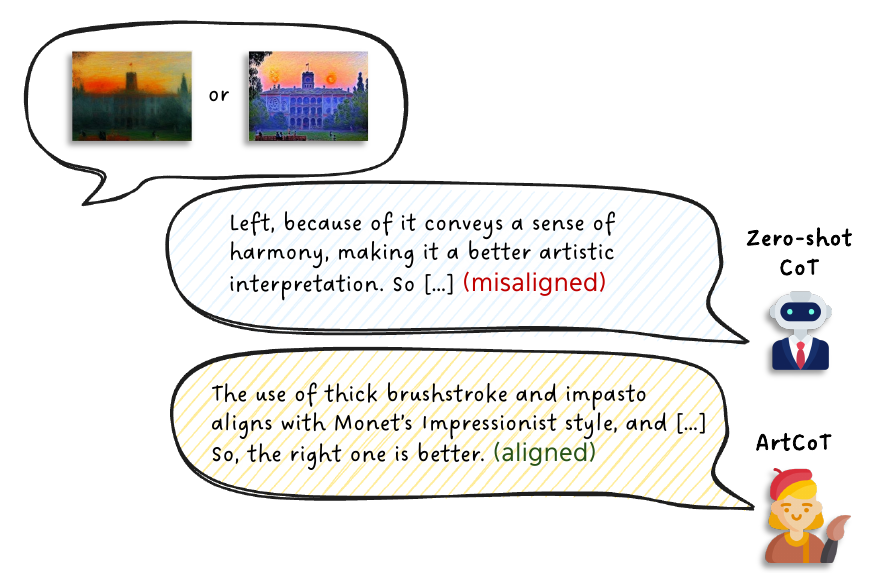}
    \caption{A minimal example of the hallucination issue. The opinionated and subjective reasoning of zero-shot CoT leads to misaligned aesthetic judgment.}
    \label{fig:hallucinate}
\end{figure}

\subsection{Zero-shot CoT Reinforces Hallucination}
\label{sec:discuss_hallucination}
During aesthetic reasoning, we identify a special pattern, where MLLMs tend to hallucinate feelings rather than to reason objectively. This tendency significantly degrades an MLLM's aesthetic judgment, leading to the counter-intuitive result where zero-shot CoT negatively affects alignment with human judgment.

Specifically, in the zero-shot CoT setting, we observe that MLLMs tend to arrive at rapid conclusions, employing subjective and less concrete language to justify their judgments. This hallucination can be further categorized into two types. The first is the \textbf{use of opinionated and subjective words}. For instance, opinion verbs (e.g., \textit{``feels''}, \textit{``senses''}) and hedge words (e.g., \textit{``appears''}, \textit{``seemingly''}) are frequently employed. These terms often indicate a lack of in-depth consideration, weak logical connection to the given task, and are less supportive for aesthetic judgment~\cite{sircello1968subjectivity}. Another pattern is the generation of \textbf{unsubstantiated artistic interpretations} that are superficial and not substantiated with facts or explanations. These hallucinations are consistently observed in misaligned reasoning traces across different MLLMs, corroborating recent findings~\cite{liu2024mind,huang2025survey}. Detailed conversation examples of ArtCoT are provided in Appendix~\ref{supp:conversation}.

\subsection{Evidence-based Reasoning Suppresses Subjectivity and Hallucination} 

In stark contrast to base prompting and zero-shot CoT, ArtCoT produces thinking traces that are objective, logical, and well-grounded with evidence. Fig.~\ref{fig:hallucinate} gives a minimal example to demonstrate this difference. To quantify this improvement, we first measure the \textit{subjectivity} of responses by calculating the frequency of subjective verbs in the lemmatized responses of MLLMs. We also follow established practices in natural language processing (NLP) by performing lexicon-based sentiment analysis using the TextBlob subjectivity score, which ranges from 0 (objective) to 1 (subjective).

For a fine-grained human evaluation of hallucinations, we analyze cases where the MLLM's preference disagreed with human judgment. We asked users to read the MLLM's reasoning and label hallucinations, which we categorize into two primary types. The first, \textbf{Factual errors}, includes incorrect image recognition (visual) and inaccurate style explanations (textual). They are also called intrinsic hallucinations~\cite{bai2024hallucination}, as they can be directly deduced from the input. Another type is \textbf{Extrinsic hallucinations}, comprising \textbf{opinionated responses} and \textbf{artistic interpretations}, which correspond to the subjective language and unsubstantiated interpretations identified in Sec.~\ref{sec:discuss_hallucination}. Users rated the severity of each of the four types of hallucinations on a scale of \verb|[0,1,2]|. The aggregated results are presented in Tab.~\ref{tab:subjective}. 

Human evaluation reveals that all models exhibit relatively less severe hallucinations in terms of visual and textual aspects. This suggests that factual inaccuracies are not the primary source of hallucinations in this context. Instead, the prevalent use of subjective language, characterized by opinionated expressions and unsubstantiated artistic interpretations, significantly impedes the ability of MLLMs to align their aesthetic judgments with those of humans. Moreover, the elevated response subjectivity observed in zero-shot CoT, as evidenced by the remarkably consistent TextBlob scores ($\approx0.45$) and the frequency of subjective words, provides further support for our finding.

The human evaluation further supports our central hypothesis that the bottleneck for MLLMs in aesthetic reasoning is not \textit{visual recognition} but \textit{reasoning elicitation}.  All models demonstrate sufficient comprehension of the input; the core challenge is preventing them from defaulting to subjective shortcuts. We illustrate this challenge is surmountable. By enforcing an evidence-based, reflective structure, ArtCoT directly mitigates subjective expression, substantially suppressing hallucinations and driving significant gains in human alignment.

\begin{figure}[]
    \centering
    \includegraphics[width=\linewidth]{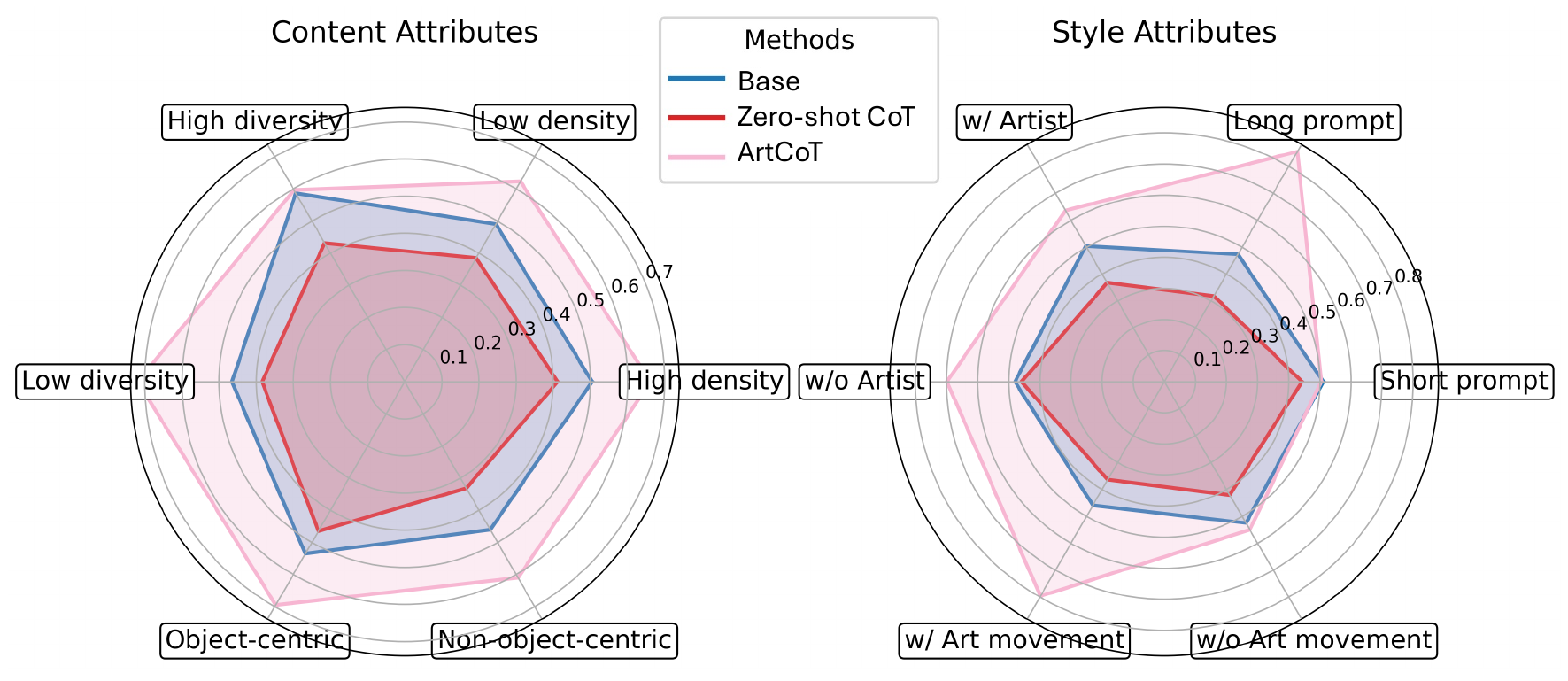}
    \caption{\textbf{Fine-grained comparison of different MLLM prompting schemes.} We show the Spearman's $\rho$ for per-instance alignment, grouped by representative attribute in FineArtBench. ArtCoT elicits aesthetic reasoning in all scenarios, especially for tasks with a specific style instruction.} 
    \label{fig:radar}
\end{figure}

\subsection{Multifaceted Aesthetic Reasoning is Versatile}
The semantic label in \textbf{FineArtBench} enables fine-grained understanding of MLLM's performance. For instance, in Fig.~\ref{fig:radar} , we visualize the alignment performance on different content image complexity and style prompt categories. The proposed ArtCoT outperforms both base and zero-shot CoT prompting in all examined sub-classes, demonstrating its versatility. 

The performance gap is most pronounced on tasks with concrete style instructions, such as those involving ``long prompt'' and prompts specifying particular ``art movement''. We posit that such specific instruction sets a more concrete objective for both the art creation and evaluation~\cite{bai2024hallucination}. Compared with obscure instructions, this specificity facilitates the MLLM to more objectively perform the two-hop reasoning, resulting in reduced hallucination and improved human alignment~\cite{huang2025survey,wei2022chain,guo2025deepseek,bai2024hallucination}.

\subsection{Ablations}
\textbf{Components of ArtCoT.}
We ablate the key components of ArtCoT, specifically the CS analyzer and the art critic, and report the results in Tab.~\ref{tab:ablate_module}. The complete ArtCoT prompt achieves the highest aesthetic alignment. Particularly, removing the art critic phase induces the most significant decline, underscoring its critical role. This provides further evidence that  reasoning is the key bottleneck.

\textbf{Input to MLLM.} We ablate input modalities and image resolutions, with results reported in Tab.~\ref{tab:ablate_input}. The full resolution achieves the best result. This is because artwork details, such as strokes, are important aesthetic factors. For input modalities, providing style information is important in per-artist alignment, while including the reference image affects per-instance alignment the most. For either setting, providing all input modalities achieves the best alignment, meaning that task-constrained 2AFC aids objective aesthetic reasoning and judgment to better align with human value.

\begin{table}[]
    \caption{\textbf{Ablation on component of ArtCoT.} We ablate the content/style analyzer and the art critic. The full two-stage aesthetics reasoning facilitates human-aligned judgment.}
    \centering
        \resizebox{0.82\columnwidth}{!}{
    \begin{tabular}{cccc}
    \toprule
         CS-analyzer & Art Critic & Per-artist $\rho$ 
         $\uparrow$& Per-task $\rho$ $\uparrow$\\\midrule
         
         \redcross &\greentick&0.630&0.532\\
         \greentick& \redcross&0.531& 0.366 \\
         \midrule
        \greentick&\greentick&\textbf{0.739}& \textbf{0.607}\\
        \bottomrule
    \end{tabular}
    }

    \label{tab:ablate_module}
\end{table}

\begin{table}[h]
    \caption{\textbf{Ablation on image resolution and source information.} We report the correlation $\rho$ (averaged from BT and Elo) of different input setups: content image, style prompt, and image sub-sampling factor.}
    \centering
    \resizebox{0.97\columnwidth}{!}{
    \begin{tabular}{ccccc}
    \toprule
         Content &Style & Resolution & Per-artist $\rho$ 
         $\uparrow$& Per-task $\rho$ $\uparrow$\\\midrule
         
         \greentick&\greentick&1/2&0.630\changeindicator{-42} &0.432\changeindicator{-44}\\
         \greentick&\greentick&1/4&0.502\changeindicator{-91}& 0.285\changeindicator{-82} \\
         \redcross&\redcross&full& 0.476\changeindicator{-100}& 0.416\changeindicator{-49}\\
         \redcross&\greentick&full&0.678\changeindicator{-23}& 0.465\changeindicator{-36}\\
         \greentick&\redcross&full&0.557\changeindicator{-69}& 0.521\changeindicator{-22}\\
         \midrule
        \greentick&\greentick& full &\textbf{0.739}& \textbf{0.607}\\
        \bottomrule
    \end{tabular}
    }
    \label{tab:ablate_input}
\end{table}

\subsection{Broader Impact}
\begin{figure}
    \centering
    \includegraphics[width=\linewidth]{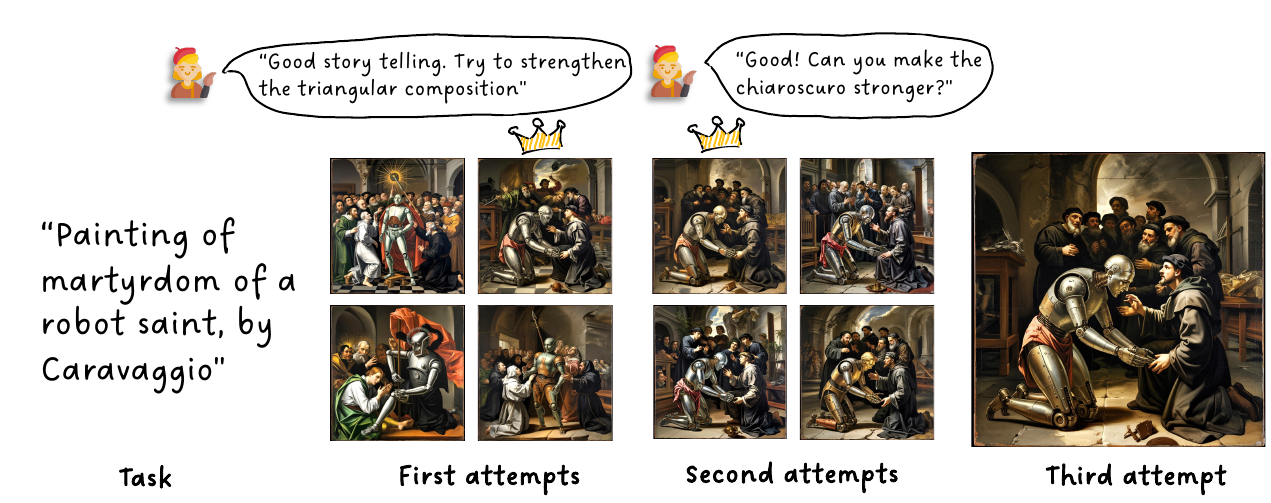}
    \caption{A minimal illustration of ArtCoT-enhanced art generation.}
    \label{fig:rlaif}
\end{figure}

We have demonstrated that evidence-based reasoning makes MLLMs human-aligned art evaluators. Many applications can benefit from this exploration, both in terms of RLAIF and HCI. Fig.~\ref{fig:rlaif} illustrates how generative art may benefit from the reward signals with aesthetic reasoners to improve the generation, utilizing a pipeline akin to ~\cite{guo2025can}. A similar pipeline may also be employed for human-centered and interactive art education.

\subsection{Limitations and Future Works}
Aesthetics, a subject studied for centuries, possesses complexities that cannot be fully addressed in a single paper. Numerous avenues for improvement exist, with the most significant challenge being the computational modeling of interperson subjectivity. One possible technical direction involves conditioning Eq.~\ref{eq:conditional_prefer} on the audience identity to enable personalized aesthetic models. A deeper understanding of the human part provides valuable insights, such as the philosophical theories and discussions on aesthetic judgment~\cite{kant2024critique,sircello1968subjectivity,hullman2023artificial}.

\section{Conclusion}
Prevailing aesthetic computing approaches are fundamentally limited, reducing rich human judgment to simplistic visual scoring. True alignment requires modeling the cognitive process behind this judgment. Models that fail to do so cannot meaningfully  align with human perception. This paper addresses this gap by pioneering the task called \textit{aesthetic reasoning} with MLLMs. In-depth analysis highlights the hallucination issues stemming from subjective opinions and unsubstantiated interpretations, leading to misaligned judgments. We further identify a promising solution through an evidence-based and objective reasoning process. Ultimately, this work provides a blueprint for developing AI that does not just recognize visual appeal, but can reason about the aesthetics in-depth—a critical paradigm shift toward creating genuinely collaborative and artistically insightful systems.

\section{Acknowledgment}
This research was supported by the Hong Kong Research Grants Council (GRF-15229423).

\bibliographystyle{ACM-Reference-Format}
\balance
\bibliography{main}

\clearpage
\begin{appendices}
    \section{Additional Experiment Details}

\subsection{Human Aesthetic Judgment Collection}\label{supp:human_detail}

\begin{figure}[!ht]
    \centering
    \includegraphics[width=\linewidth]{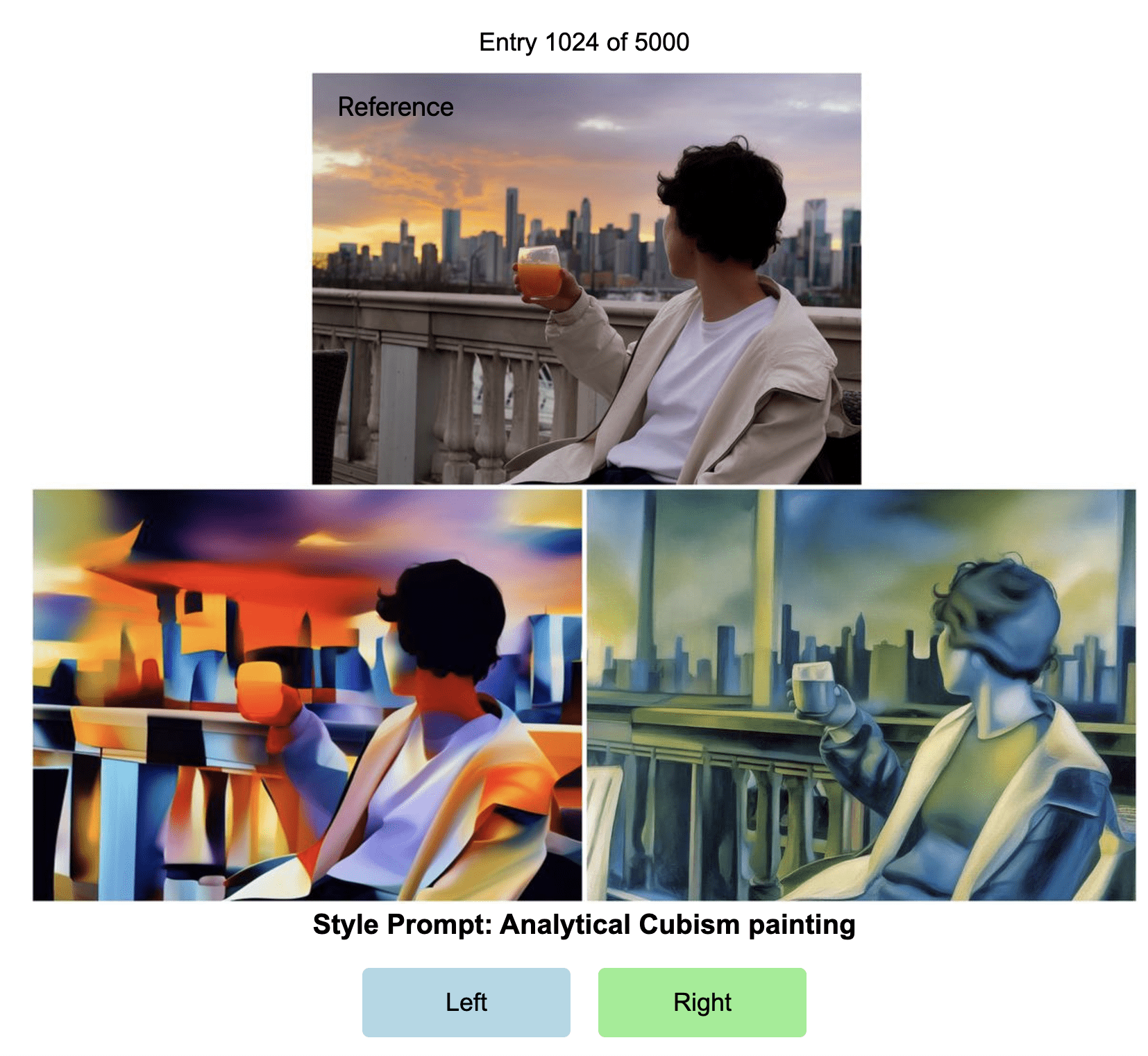}
    \caption{\textbf{UI for 2AFC annotation.} We present user with the source image (top), 2AFC (middle) and style prompt (bottom). The user is required to choose the preferred one by clicking on the ``left" or ``right" button.}
    \label{fig:ui}
\end{figure}

\begin{figure*}[!ht]
    \centering
    \includegraphics[width=\linewidth]{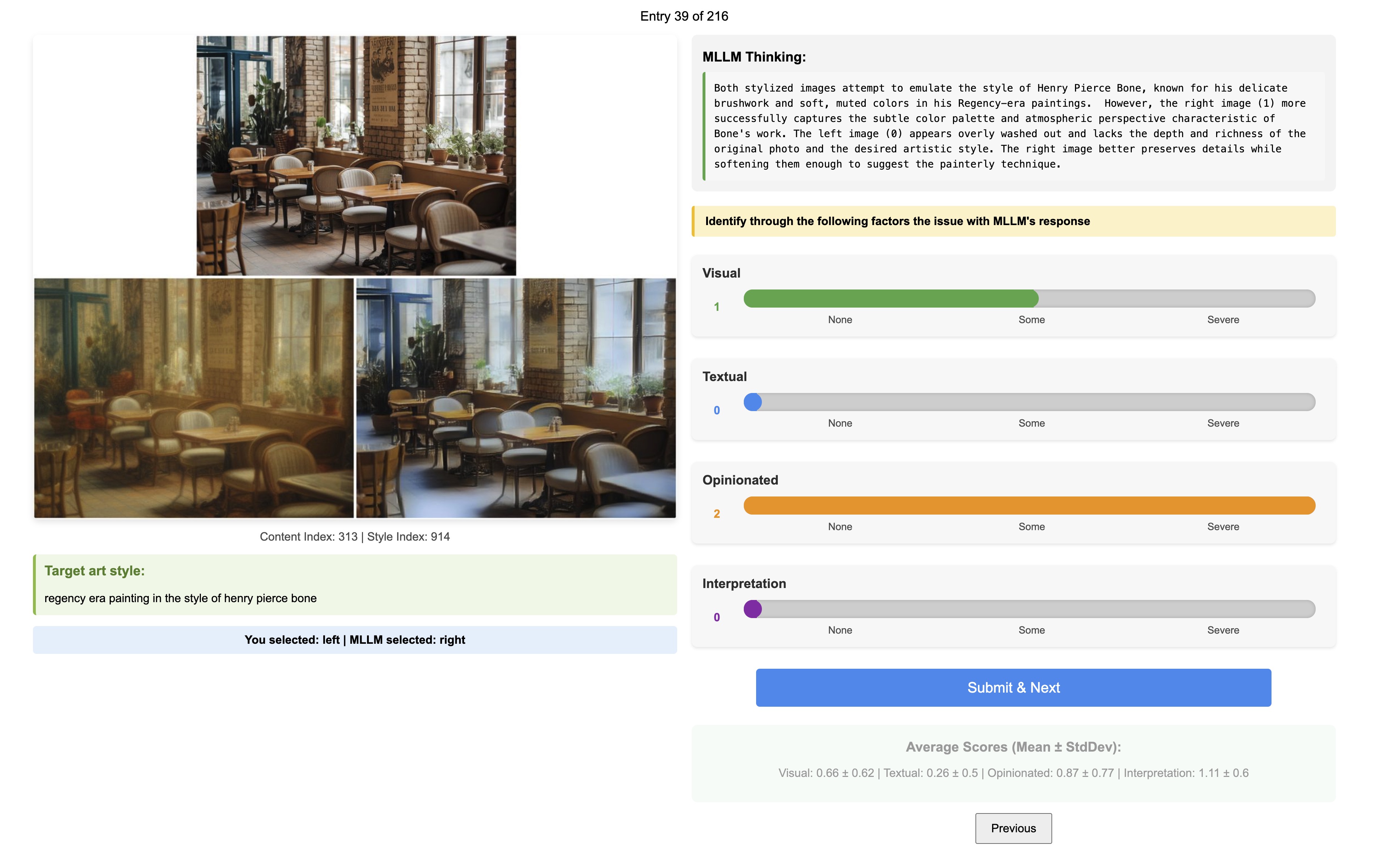}
    \caption{UI for hallucination annotation. Users are presented with the original 2AFC task and the response of MLLM, when the answer are not aligned. User shall drag discrete-valued slider to determine the hallucination strength from four perspectives.}
    \label{fig:ui_hallucination}
\end{figure*}

\textbf{User Interface}
For the main judgment collection in Sec.~\ref{sec:human_model}, we develop a web-based application to collect responses from human-expert annotators. We present users with the 2AFC question and reference content and style. The content is presented as a reference image, while we present the style as a prompt. The name of the stylization model is tracked but not visible to the user during annotation. The screenshot of the user interface (UI) can be found in Fig.~\ref{fig:ui}. 

For the hallucination labeling, we also designed a UI, visualized in Fig.~\ref{fig:ui_hallucination}. We record the user's response during the 2AFC annotation and present them with the misaligned comparisons. The user shall read the MLLM's response (the prompting method and the model name is not visible) and determine the hallucination issue from the four aspects. The image on left will be enlarged on mouse hover.

The annotation tools will be open-sourced to benefit future research.



\subsection{The Sampling algorithm} We formally describe the proposed degree-uniform sub-graph sampling algorithm in Algo.~\ref{algo: subsample}. \label{append: algo}

\begin{algorithm}[]

\resizebox{\linewidth}{!}{%
\begin{minipage}{\linewidth}
\scriptsize  
\caption{Sample a Connected Subgraph with Uniform Degree Distribution}
\label{algo: subsample}
\KwIn{
   $G = (V, E)$ \tcp*{Complete graph} \\
   $n$ \tcp*{Number of edges, $|V| - 1 \leq n \leq \frac{|V|(|V| -1)}{2}$} \\
   $\text{RNG}$ \tcp*{Random Number Generator}
}
\KwOut{
   $E' \subseteq E$ \tcp*{Subsampled edge set forming a connected subgraph}
}

\textbf{1. Validate Inputs:}\\
\If{$|V| < 1$ \textbf{or} $n < |V| - 1$ \textbf{or} $n > \frac{|V|(|V| -1)}{2}$}{
   \textbf{Error:} Invalid input parameters.\;
}

\textbf{2. Generate Spanning Tree:}\\
$E_T \gets$ Kruskal's MST($G$, $\text{RNG}$)\;

\textbf{3. Initialize Subgraph and Degrees:}\\
$E' \gets E_T$\;
Initialize $d(v) = 0,\ \forall v \in V$\;
\ForEach{$e = (u, v) \in E_T$}{
   $d(u) \gets d(u) + 1$\;
   $d(v) \gets d(v) + 1$\;
}

\textbf{4. Add Remaining Edges:}\\
$m \gets n - |E_T|$\;
\While{$m > 0$}{
   \textbf{a. Identify Best Candidate Edges:}\\
   \[
       BestEdges \gets \left\{ e \in E \setminus E' \ \bigg| \ 
       \min \max\{d(u)+1, d(v)+1\} \right\}
   \]
   \[
       BestEdges \gets \left\{ e \in BestEdges \ \bigg| \ 
       \min \sum\{d(u)+1, d(v)+1\} \right\}
   \]
   
   \textbf{b. Select and Add an Edge:}\\
   $e^* \gets$ Randomly select from $BestEdges$ using $\text{RNG}$\;
   $E' \gets E' \cup \{e^*\}$\;
   
   \textbf{c. Update Degrees and Counter:}\\
   \ForEach{$v \in e^*$}{
       $d(v) \gets d(v) + 1$\;
   }
   $m \gets m - 1$\;
}

\Return{$E'$}\;
\end{minipage}%
}
\end{algorithm}


\subsection{Detail on FineArtBench}\label{supp:dataset_detail}

We construct FineArtBench by harvesting from existing open-source datasets with the help of MLLMs. 

For the \textbf{Content} category, 50\% of the content images are generated using the Ideogram-v1 text-to-image (T2I) diffusion model with diverse prompts produced by GPT-4. The remaining 50\% are randomly sampled from the SA-1B~\cite{kirillov2023segment} and MS-COCO~\cite{lin2014microsoft} datasets, with captions generated by Gemini-v1.5 pro~\cite{team2023gemini}. Images from SA-1B are downsampled by a factor of two, while those from MS-COCO retain their original resolution. Overall, all images have an average height and width of 895.7 and 811.9 pixels, respectively. The 5th percentiles for height and width are 480 and 427 pixels, respectively, and the 95th percentiles for both dimensions are 1248 pixels.

To synthesize fine-grained attribute annotations for the content images and their associated prompts, we first employ Gemini v1.5-pro to automate the annotation process based on a predefined attribute set. The \textbf{Style} subset of FineArtBench is derived from WikiArt~\cite{wikiart} and DiffusionDB~\cite{wang2022diffusiondb}. The process involves two main steps:

\begin{enumerate}
    \item \textbf{WikiArt Processing}: We extract keywords from WikiArt, focusing on specifications such as art movement, artist, and genre. These base keywords are then expanded and combined using GPT-4~\cite{openai2023gpt} to create a diverse set of style descriptors.
    
    \item \textbf{DiffusionDB Processing}: Given the high noise level in DiffusionDB annotations, we utilize an MLLM to preprocess the text prompts within the dataset. Specifically, we extract the style descriptions from each prompt and merge those with similar style specifications. Subsequently, we subsample from these processed style prompts to ensure quality and diversity.
\end{enumerate}

For both the WikiArt and DiffusionDB prompts, style reference images are generated using StableDiffusion-v2~\cite{rombach2022high}, with all images standardized to a resolution of 512$\times$512 pixels. Similar to content annotations, we employ MLLMs to annotate the attributes of style prompts.

At least two human annotators manually validate all annotations to ensure quality. The actual number of sampled content and style images is summarized in Table~\ref{tab:combined_sources}. We will open-source FineArtBench to facilitate reproducibility and support future research endeavors.


\begin{table}[!ht]
    \centering
    \caption{\textbf{Content and Style Sources} FineArtBench is built from diverse sources to eliminate bias.}
    \label{tab:combined_sources}
    \resizebox{\linewidth}{!}{
    \begin{tabular}{l *{5}{c}}
        \toprule
        & \multicolumn{3}{c}{\textbf{Content }} & \multicolumn{2}{c}{\textbf{Style}} \\
        \cmidrule(lr){2-4} \cmidrule(lr){5-6}
        \textbf{Source} & \textbf{Generated} & \textbf{MS-COCO} & \textbf{SA-1B} & \textbf{WikiArt} & \textbf{DiffusionDB} \\
        \midrule
        \textbf{Number} & 500 & 250 & 250 & 764 & 236 \\
        \bottomrule
    \end{tabular}}
\end{table}

\subsection{Prompt Design}\label{supp:prompt_detail}

We summarize the full prompt of base prompt, zero-shot CoT prompting and the proposed ArtCoT prompt in Tab.~\ref{tab:MLLM_prompt}. We did not attempt to optimize the prompt design of ArtCoT as it just act as a strong baseline method to identify the key issue and inspire future works.


\begin{table*}[]
\centering
\caption{\textbf{Template for different prompting methods.} [STYLE] stands for placeholder for the style prompt, and [IMAGE] stands for placeholder for image tokens.}
\begin{tabular}{p{0.18\textwidth} p{0.18\textwidth} p{0.18\textwidth} p{0.18\textwidth} p{0.18\textwidth}}
\toprule
\multicolumn{1}{c}{\multirow{2}{*}{Base Prompt}} & 
\multicolumn{1}{c}{\multirow{2}{*}{Zero-Shot CoT}} & 
\multicolumn{3}{c}{ArtCoT} \\ 
\multicolumn{1}{c}{} & 
\multicolumn{1}{c}{} & 
\multicolumn{1}{c}{CS Analyzer} & 
\multicolumn{1}{c}{Art Critic} & 
\multicolumn{1}{c}{Summarizer}  \\ \midrule 
 `[IMAGE]` You are an expert in fine art. A source image (top) and two different stylized images (bottom) in the style of `[STYLE]`  are presented to you. 
        Consider both the content and style, which stylized image is better in terms of overall aesthetic quality as an artwork? Return your decision in a Python Dict, ['winner':int]. `0` means the left is better while `1` means the right is better. Do not answer any other things. 
        & `[IMAGE]`
        \{"request": "You are an expert in fine art. A source image (top) and two different stylized images (bottom) in the style of `[STYLE]` are presented to you. Consider both the content preservation and style fidelity, which stylized image is better in terms of overall aesthetic quality as an artwork?". Return the reason and your decision in short in format of a Python Dict {{ 'thinking':str, 'winner':int}}. `0` means the left is better while `1` means the right is better.",
    "response": "\{'thinking': ' Let's' think step by step,
    & `[IMAGE]`
     You are an expert in fine art. A source image (top) Two stylized images (bottom left and bottom right) in the style of `[STYLE]` are presented to you. Compare the content preservation and style fidelity of the two images, which one is better. Return your answer in a Python Dict, ['style\_reason':str, 'content\_reason':str, 'style\_winner':int, 'content\_winner':int]. 
     
     `0` means the left is better while `1` means the right is better. Do not include any other string in your response.
        
        & `[IMAGE]`
 Take a closer look at the two stylized images at the bottom in the style of `[STYLE]`. As an expert in art, do you agree with above analysis? Compare and consider the following questions.
What visual features is essential for the style of `[STYLE]`? Is the content at top well-preserved in the specific art style? Is there any artifact, distortion or inharmonious color patterns in either painting?     Return your answer in a Python Dict, [reflection':str].
        & `[IMAGE]` 
        Now we summarize. Based on above analysis and reflection, which stylized image at the bottom is better in terms of overall aesthetic quality as an **painting of the original content (top) in another style**? Return your answer in a Python Dict, ['winner':int]. 
        `0` means the left is better while `1` means the right is better. Do not include any other string in your response.

\\
    \bottomrule                                          
\end{tabular}%

\label{tab:MLLM_prompt}
\end{table*}
\section{Further Discussions}

\subsection{Limitation}
Despite the progress we made in this paper, there still exists a wide range of possible venue for future study. We would like to provide a more detailed discussion of them here.

\textbf{Implementation.} We mainly test our experiment with the publicly available and mainstream MLLMs. First, the main objective of this paper is to understand how CoT can be utilized for aesthetic reasoning. This reasoning is in zero-shot and not designed to be model-specific. As evidenced in our experiment, the performance gain and hallucination suppression are already consistent and statistically significant. Second, due to the computational efficiency of MLLMs' inference, we prototyped with small-scale models like LLaVa and LLaVa-Next. In practice, we find these small-scale models struggle with instruction following. Future work could investigate the performance of ArtCoT with larger, more capable open-source MLLMs.

\textbf{Art Styles.} This paper primarily deals with non-conceptual art, where aesthetic quality is more closely related to visual features (though not solely based on them). This is a key assumption of ArtCoT and FineArtBench, which cover much of mainstream and fine art. However, we acknowledge that avant-garde styles, particularly conceptual art, present different challenges. The value of conceptual art relies heavily on interpretation, which may be conveyed through various media and is often more subjective. A limitation of our work is that ArtCoT, focusing on visual features and evidence-based reasoning, may not be directly applicable to evaluating conceptual art. Future research could explore incorporating contextual information and interpretive frameworks into the aesthetic evaluation process.


\section{Example Conversations}
\label{supp:conversation}
We present several qualitative comparisons of MLLM's response generated using different prompts in the following pages. Specifically, we show the input images to Google Gemini and its raw output from different reasoning methods. We highlight the subjective words (yellow background) and the unsubstantiated interpretation of zero-shot CoT. Compared with zero-shot CoT, the responses from ArtCoT are much more detailed and contain less hallucination. 
\begin{figure*}
    \centering
    \includegraphics[width=\linewidth]{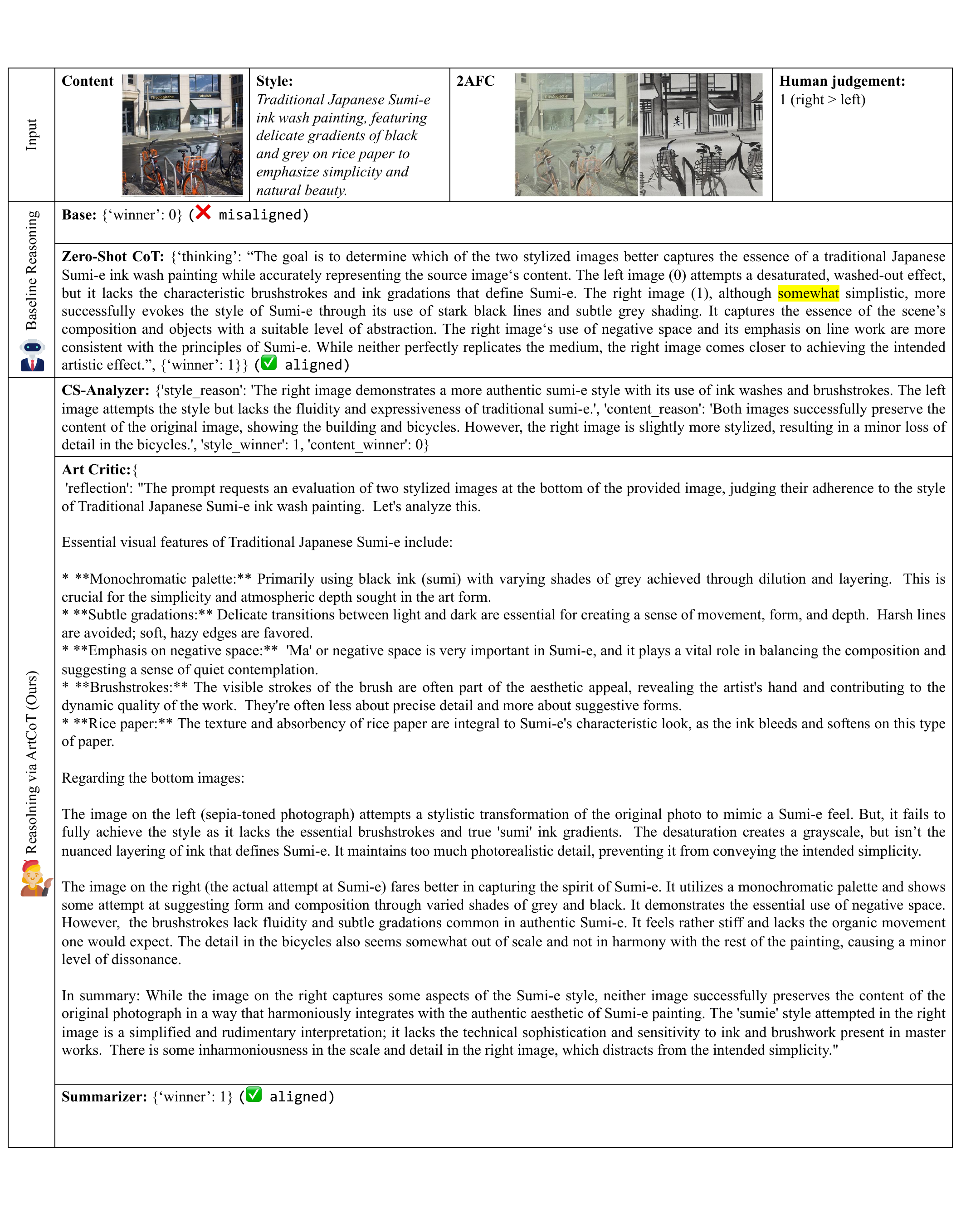}
    \label{fig:examples1}
\end{figure*}

\begin{figure*}
    \centering
    \includegraphics[width=\linewidth]{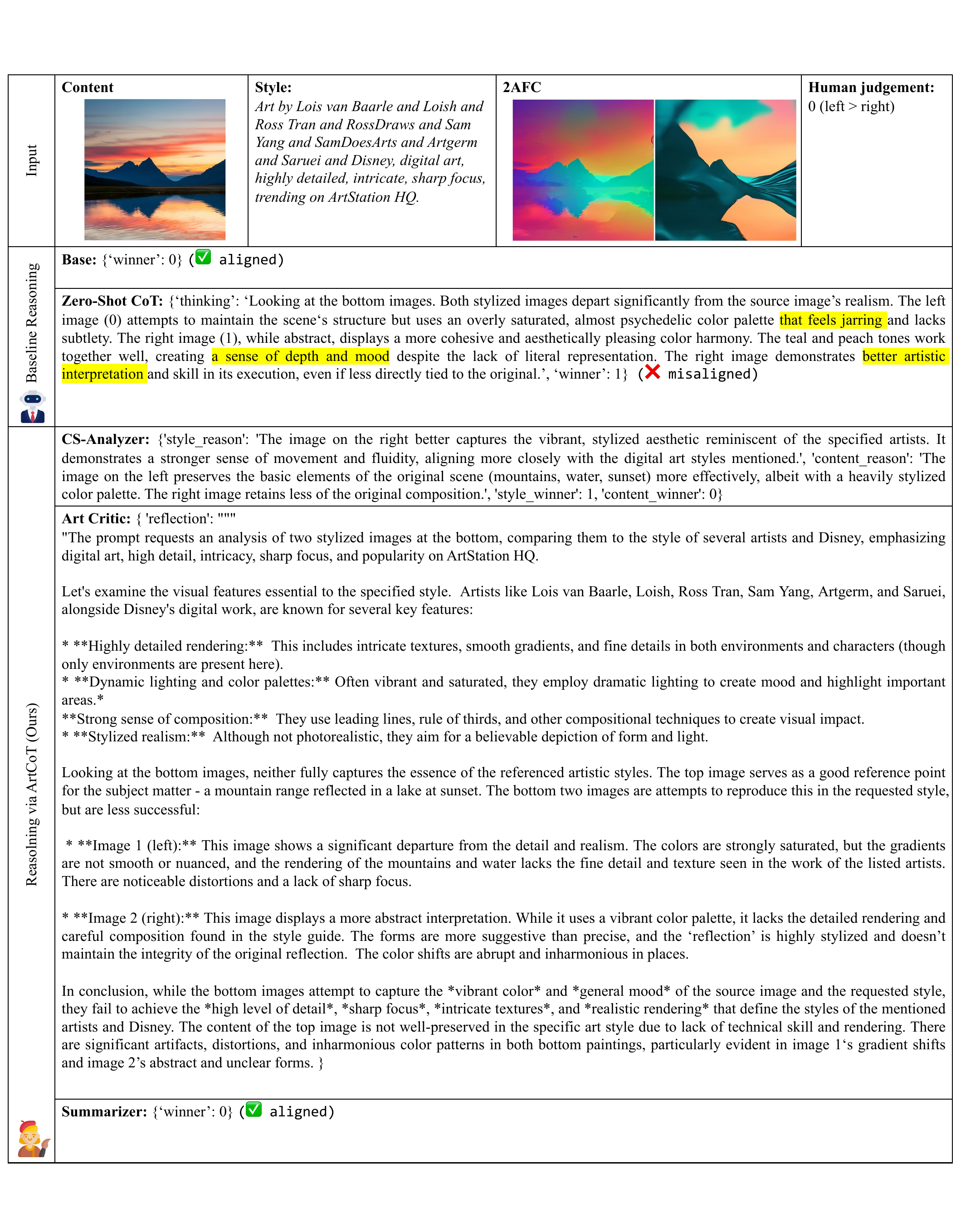}
    \label{fig:examples2}
\end{figure*}

\begin{figure*}
    \centering
    \includegraphics[width=\linewidth]{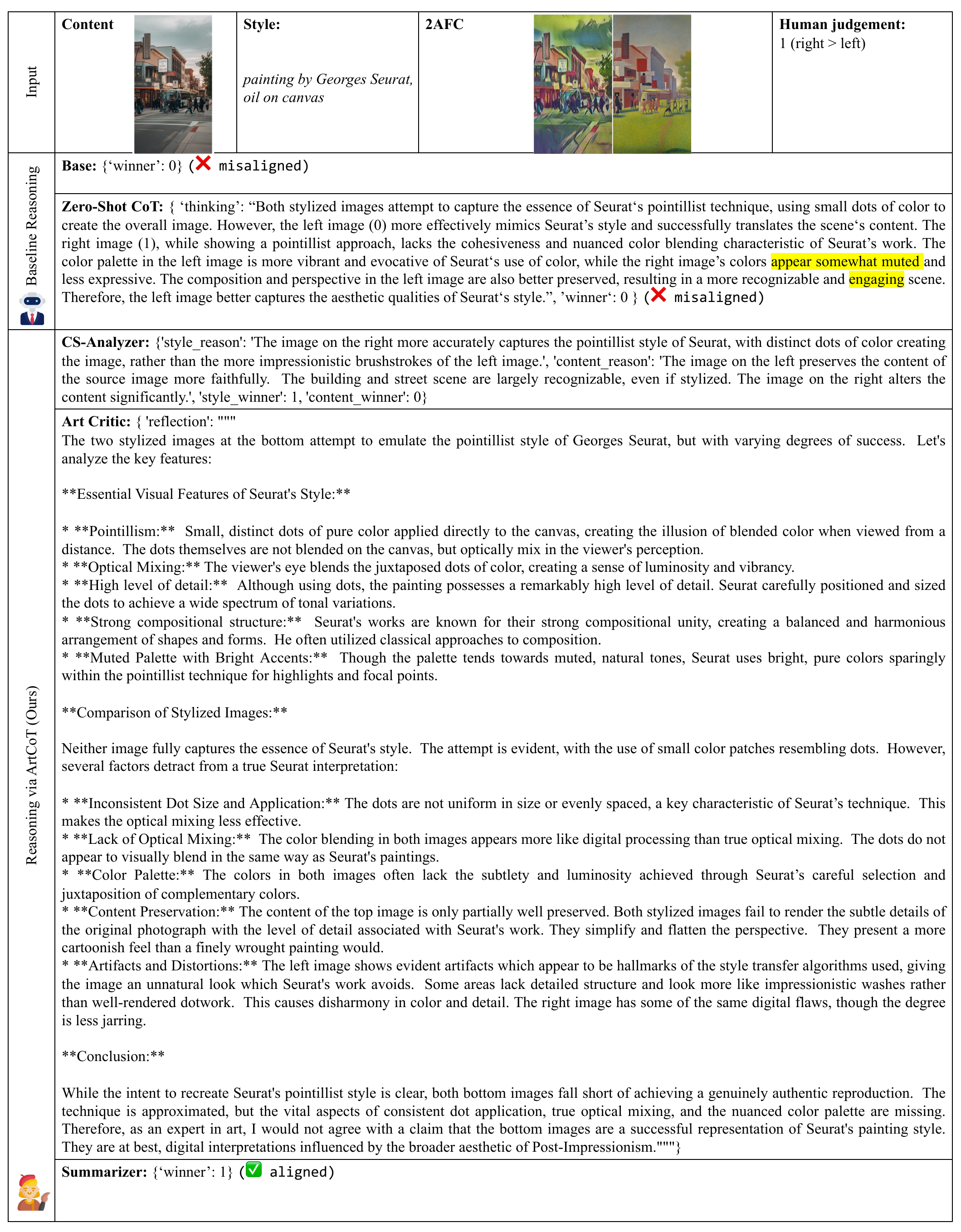}
    \label{fig:examples3}
\end{figure*}

\begin{figure*}
    \centering
    \includegraphics[width=\linewidth]{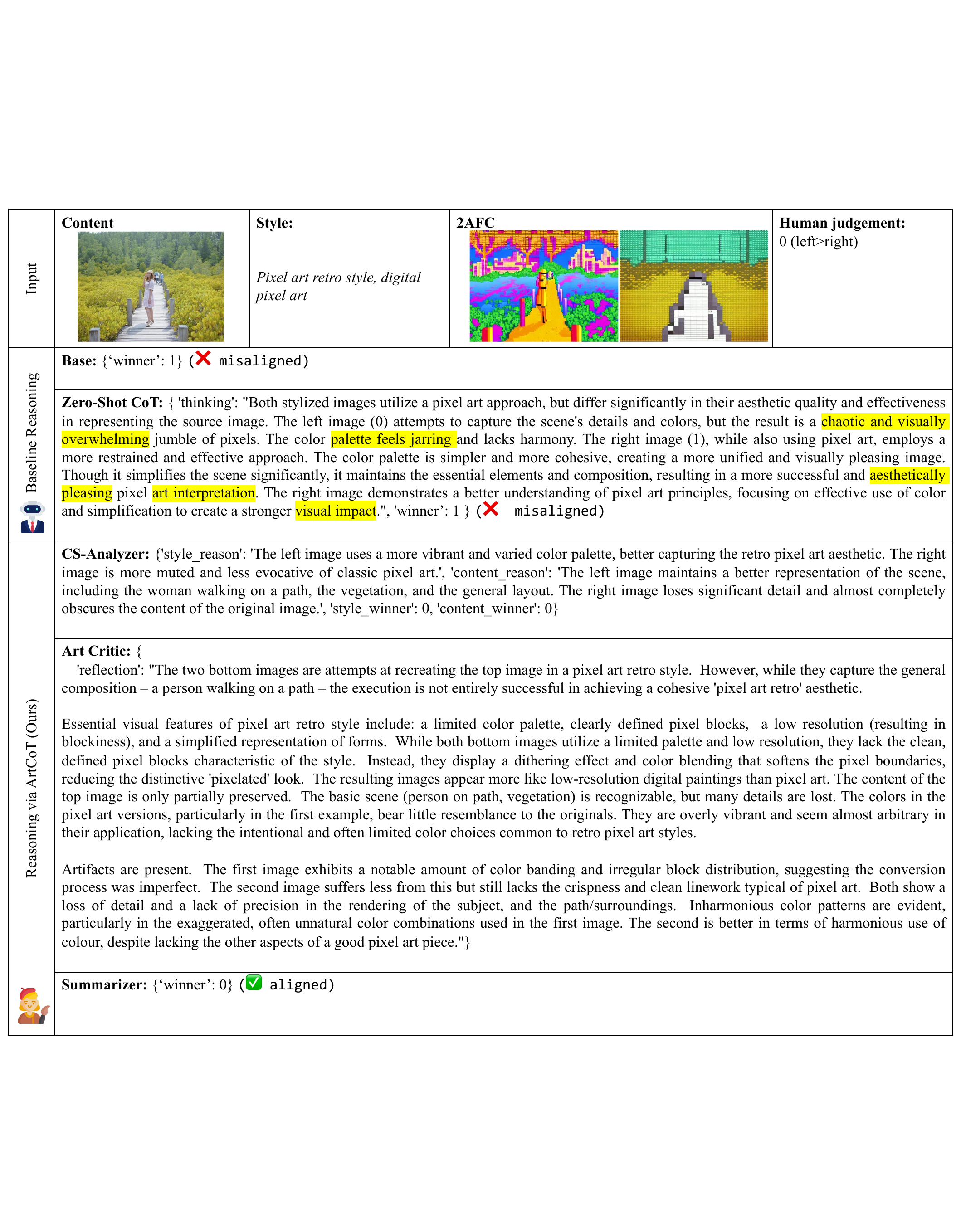}
    \label{fig:examples4}
\end{figure*}

\begin{figure*}
    \centering
    \includegraphics[width=\linewidth]{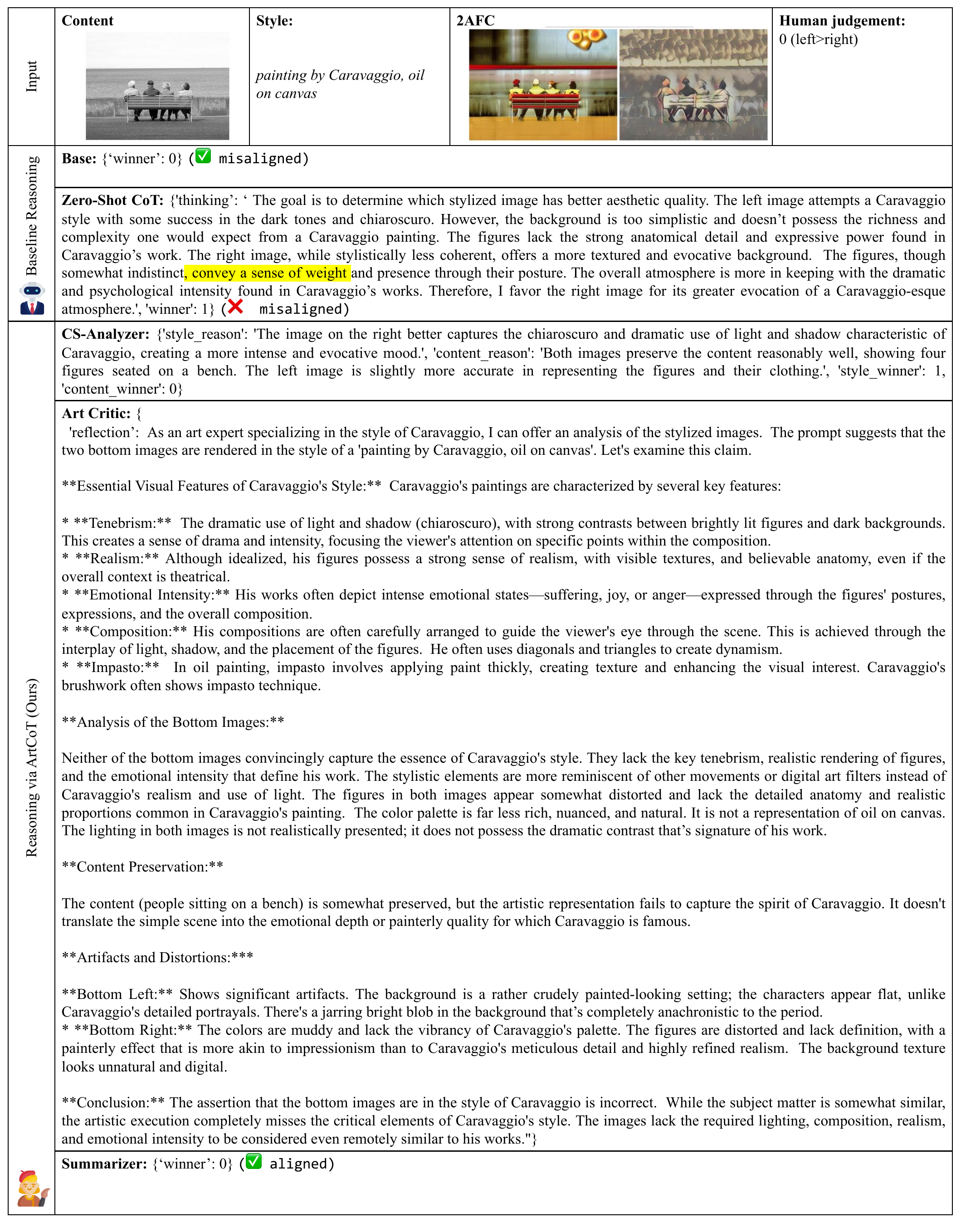}
    \label{fig:examples5}
\end{figure*}

\begin{figure*}
    \centering
    \includegraphics[width=\linewidth]{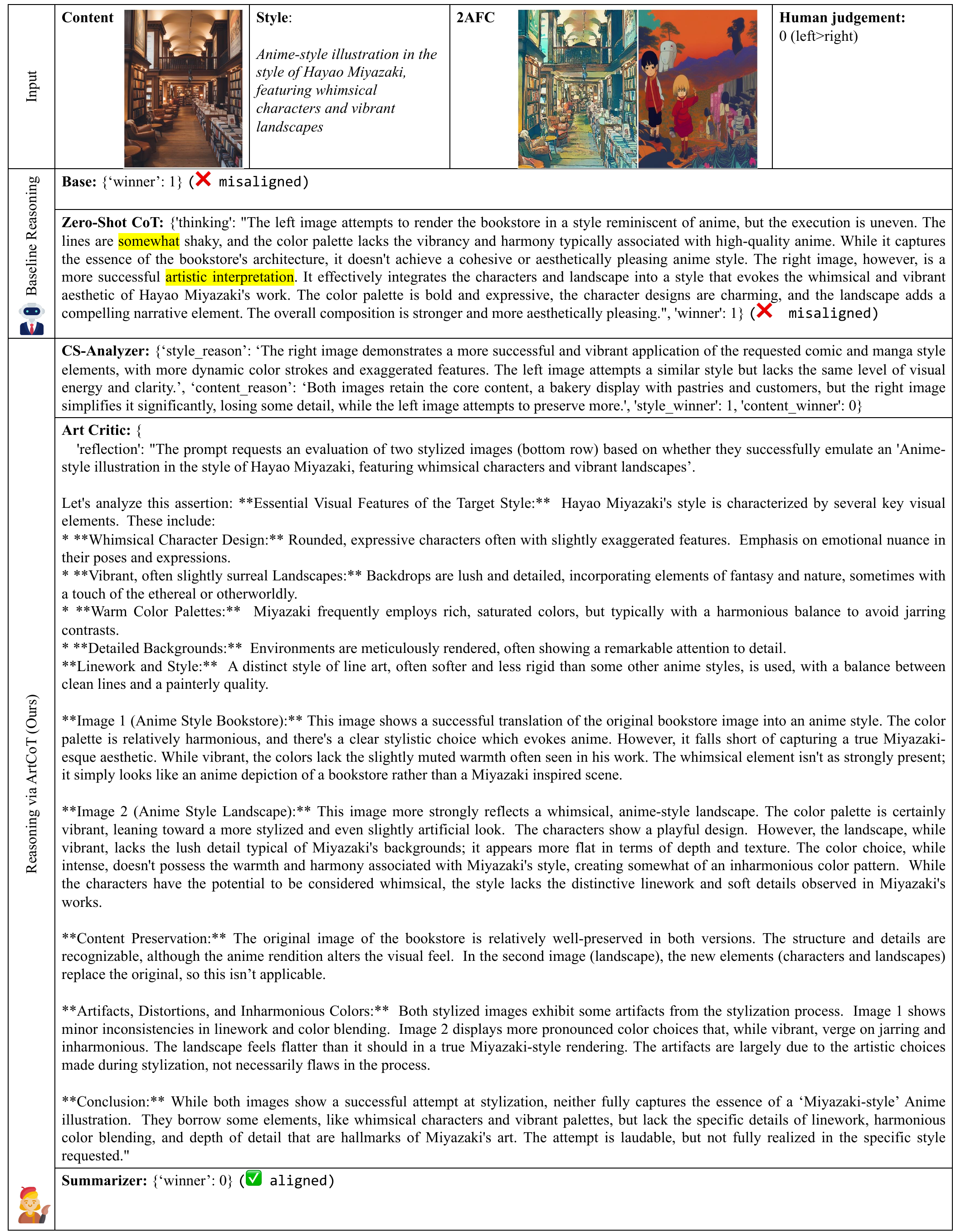}
    \label{fig:examples6}
\end{figure*}

\begin{figure*}
    \centering
    \includegraphics[width=\linewidth]{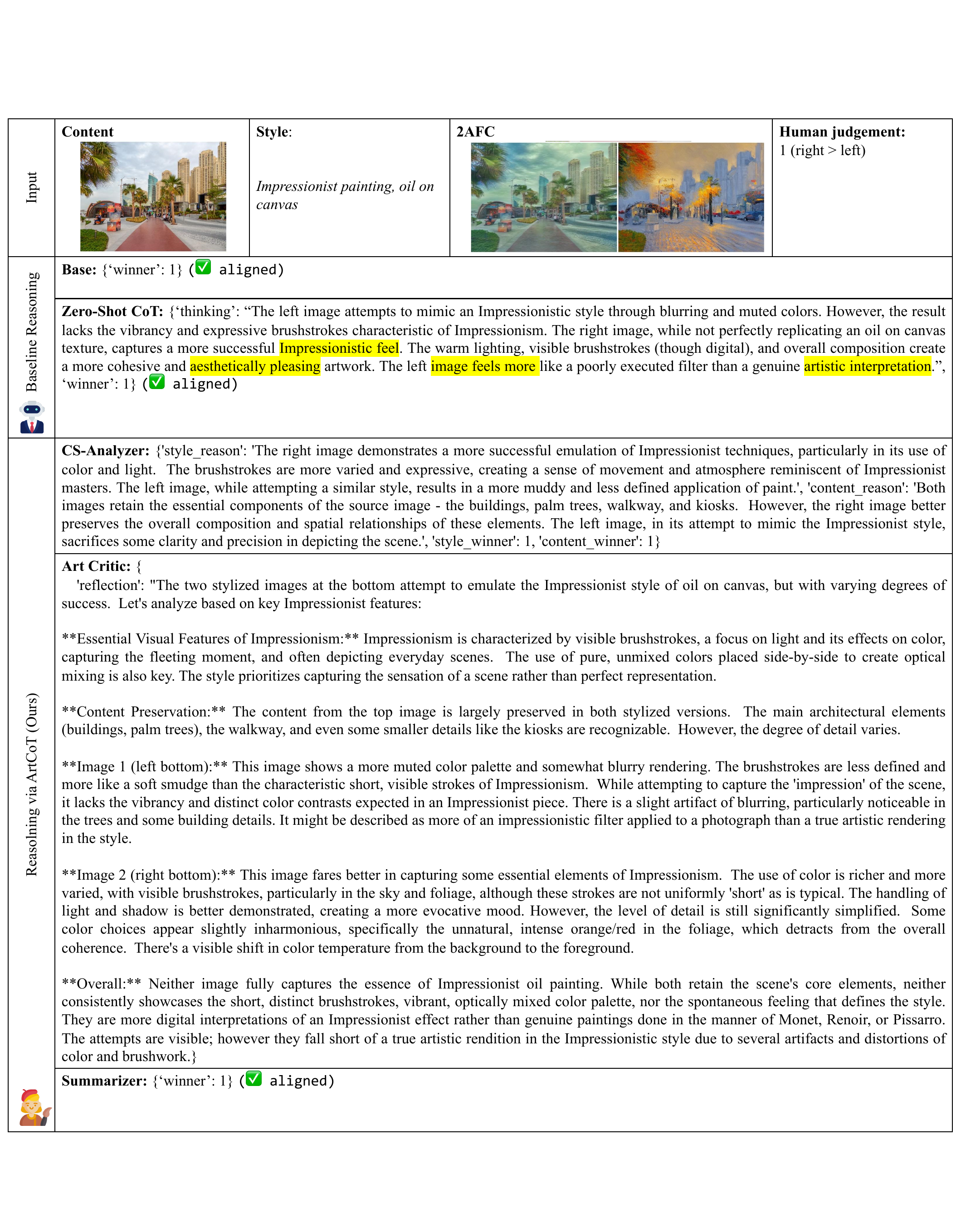}
    \label{fig:examples7}
\end{figure*}

\begin{figure*}
    \centering
    \includegraphics[width=\linewidth]{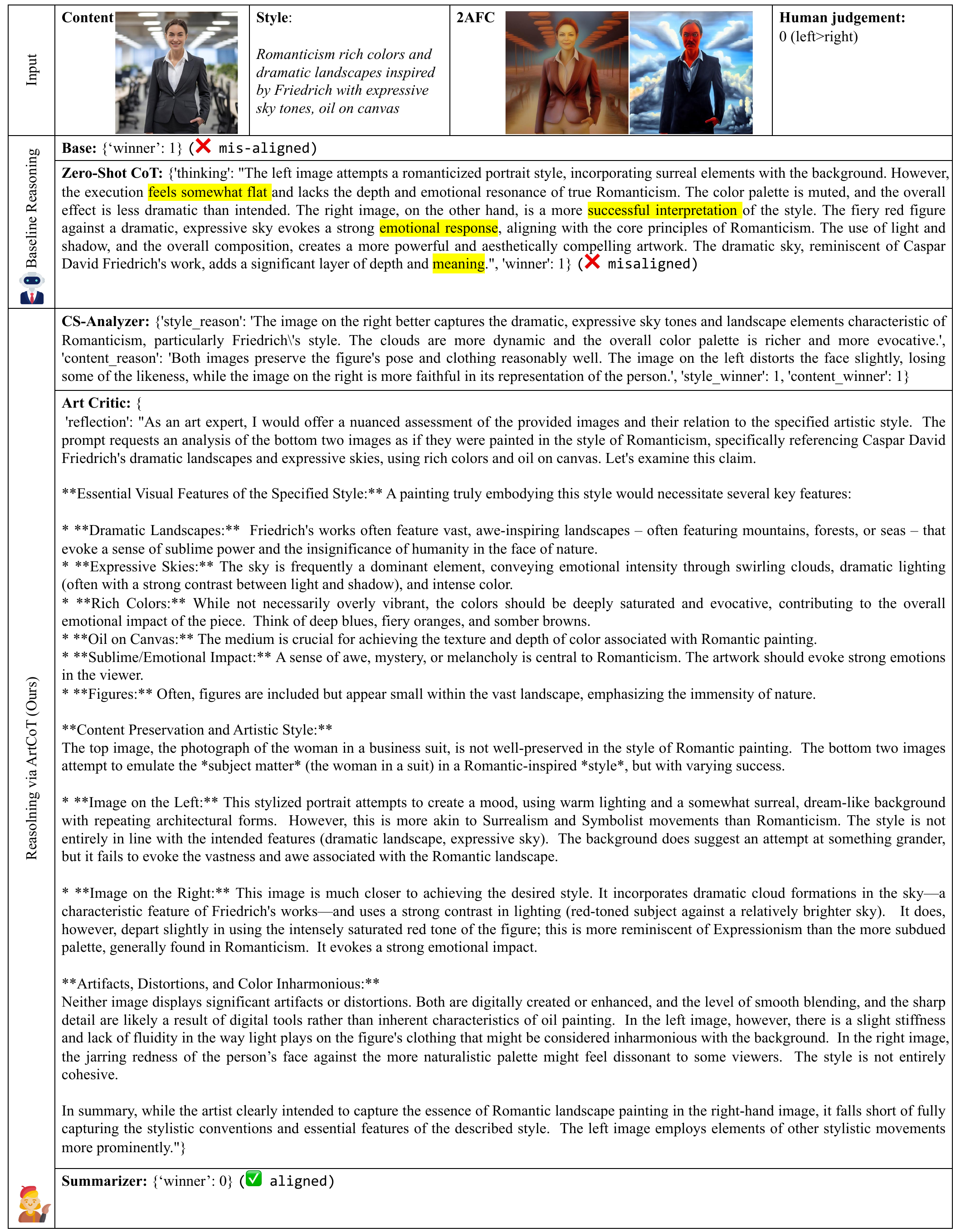}
    \label{fig:examples8}
\end{figure*}

\begin{figure*}
    \centering
    \includegraphics[width=\linewidth]{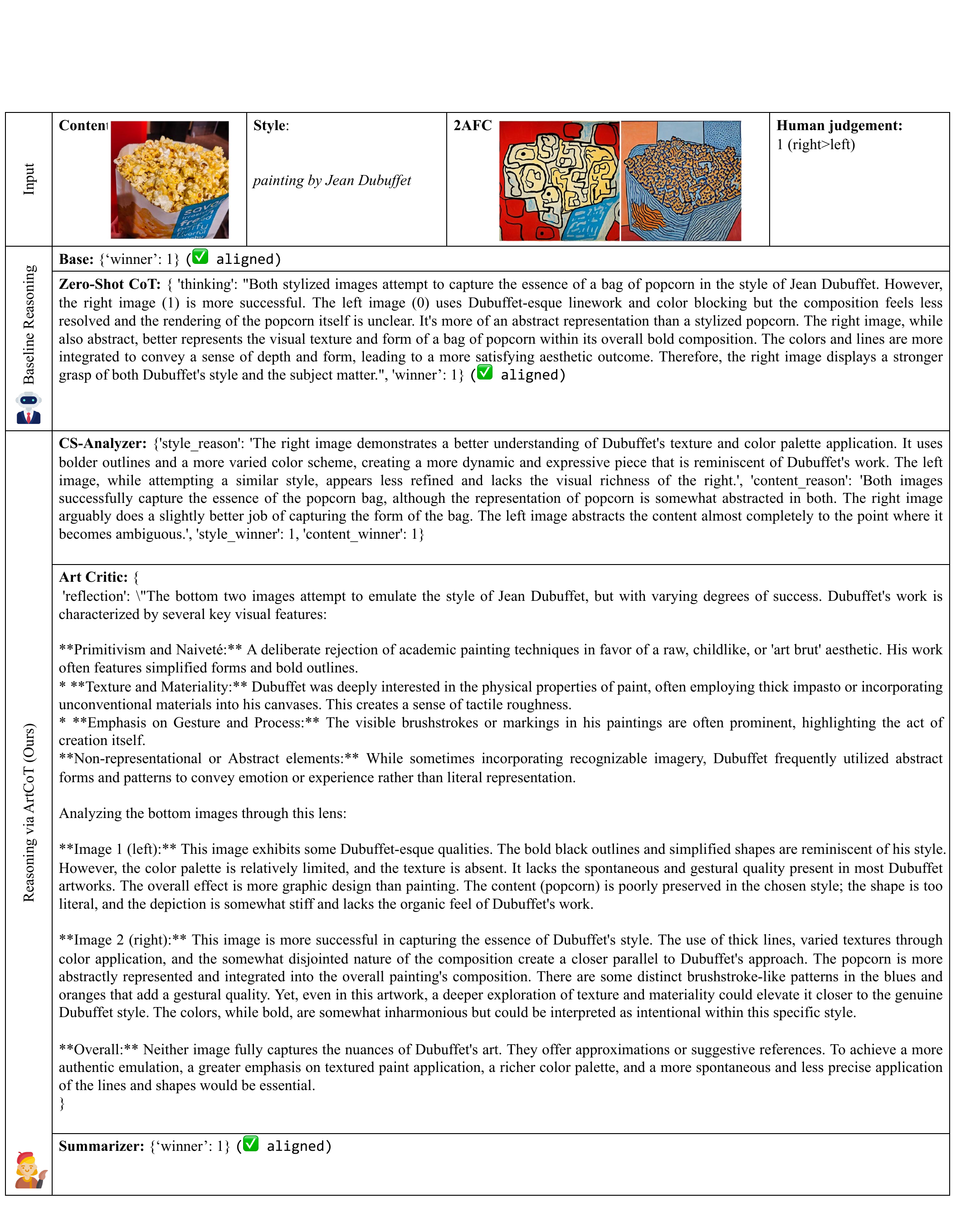}
    \label{fig:examples9}
\end{figure*}

\begin{figure*}
    \centering
    \includegraphics[width=\linewidth]{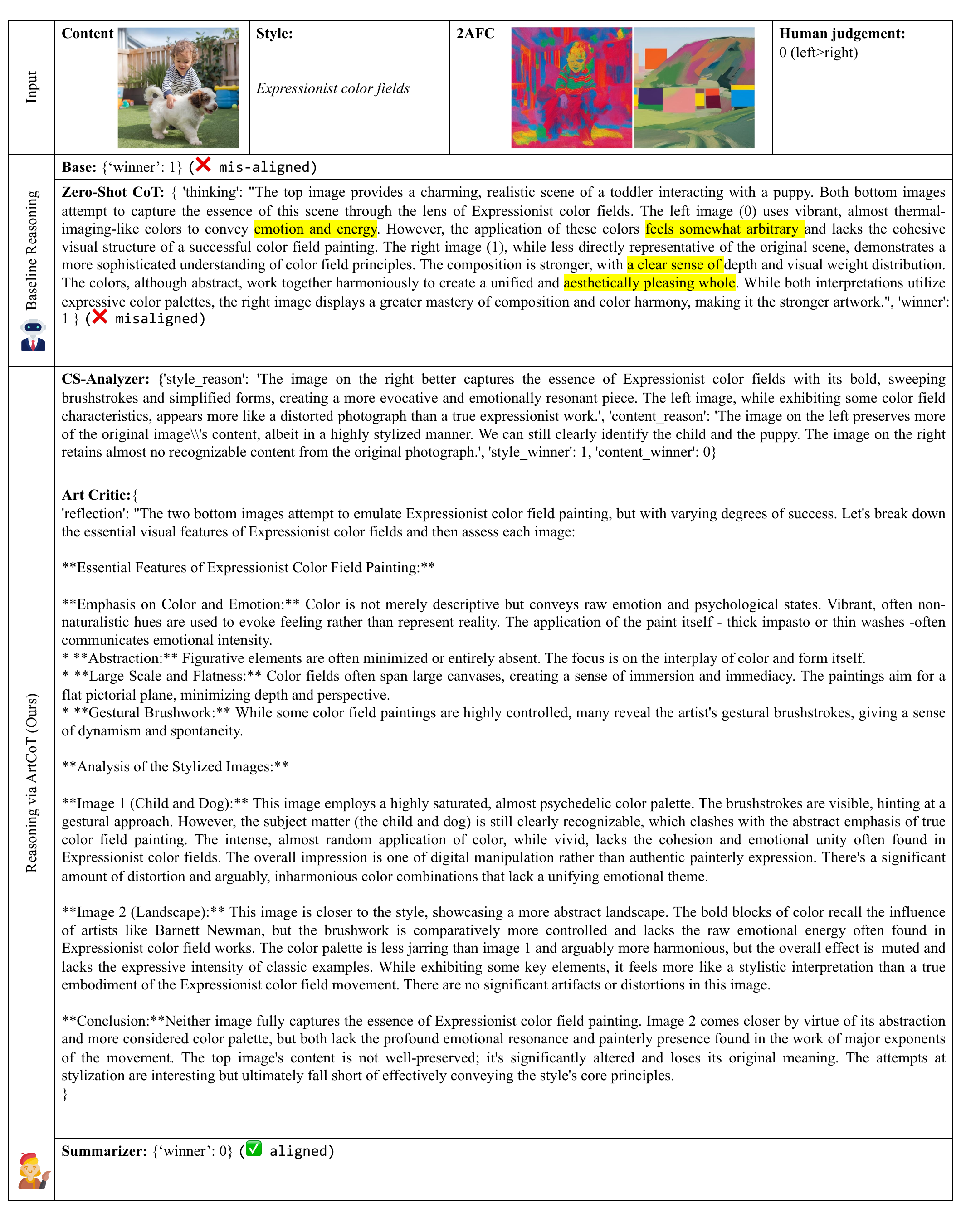}
    \label{fig:examples10}
\end{figure*}

\section{Samples of FineArtBench}
\begin{figure*}
    \centering
    \includegraphics[width=0.9\linewidth]{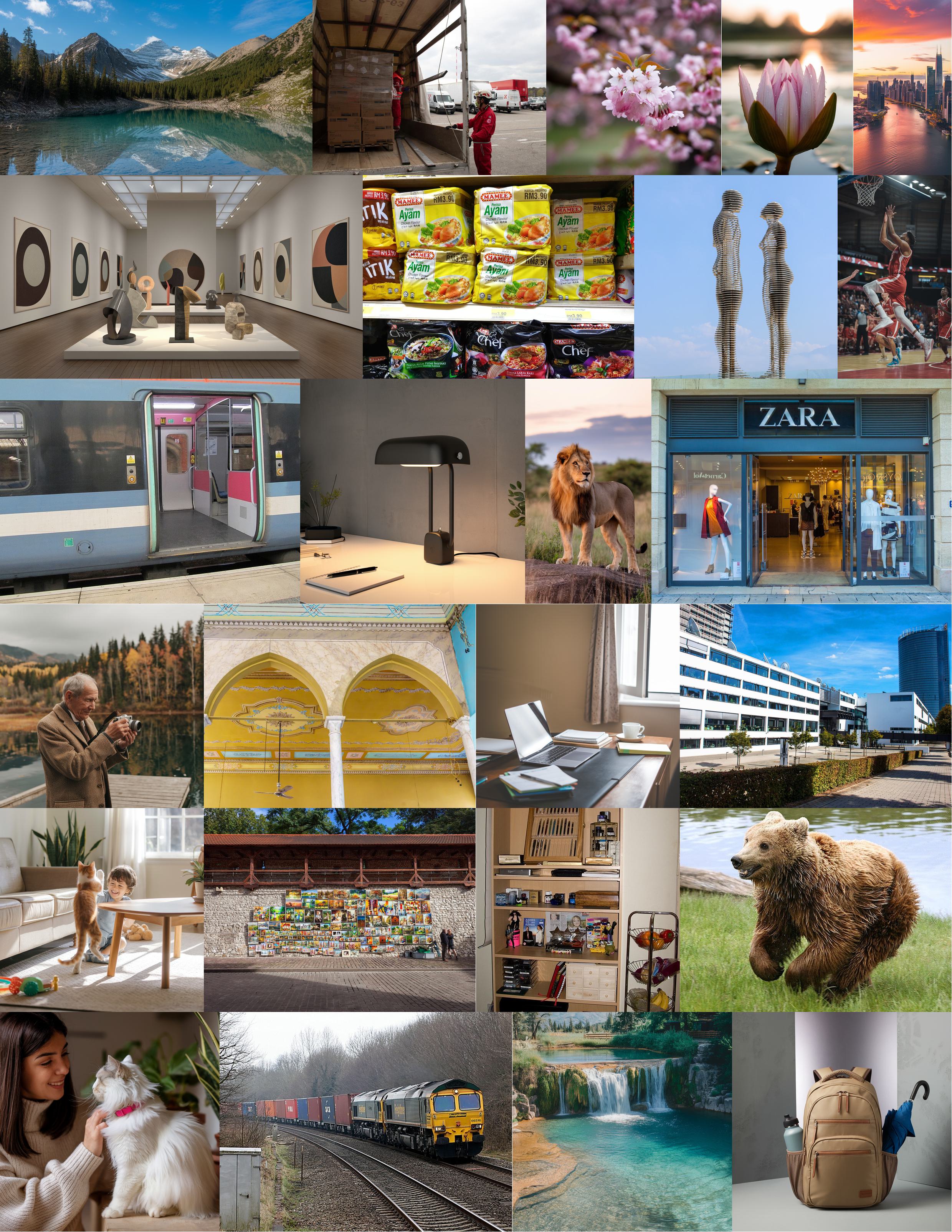}
    \caption{Random sample of content images from FineArtBench}
    \label{fig:content_example}
\end{figure*}

Examples of content image and style prompt is in Fig.~\ref{fig:content_example}, Tab.~\ref{tab: example_style}, respectively

\begin{table*}

\centering
\caption{Examples of style prompt}
\label{tab: example_style}
\begin{tabular}{>{\raggedright\arraybackslash}p{0.95\textwidth}}
\toprule
\textbf{Randomly Sampled Style Prompts} \\
\midrule
Bauhaus geometric abstraction with primary colors and clean lines \newline
The Blue Boy by Thomas Gainsborough \newline
Alphonse Mucha, Vincent van Gogh, Egon Schiele-inspired artwork with expressive color palettes and dynamic brush strokes \newline
Cafe Terrace at Night by Vincent van Gogh \newline
painting by Rumiko Takahashi, manga ink on paper \newline
painting by Donato Giancola \newline
Symbolist mystical and allegorical imagery inspired by Klimt, oil on canvas with intricate patterns \newline
Low-poly 3D model, 3D rendering \newline
guido crepax painting \newline
epic, megadeth cover, aesthetically inspired by beksinski and dan mumford, trending on artstation, art by greg rutkowski, 8k \newline
Imagery rich in symbolic and allegorical content \newline
Surrealist dream-like elements reminiscent of Dalí \newline
Cubist fragmented forms, oil on canvas \newline
Girl with a Pearl Earring, oil on canvas, Baroque portrait renowned for its enigmatic expression and masterful lighting \newline
Las Meninas \newline
Thomas Gainsborough's 'The Blue Boy', a Rococo oil on canvas portrait known for its elegant attire and striking blue palette. \newline
salvador dali painting \newline
painting by Nanna Ditzel, mixed media \newline
The Scream \newline
painting by Pierre-Auguste Renoir, oil on canvas \newline
in the style of Honor C. Appleton \newline
painting by Grandma Moses \newline
Anime-style illustration in the style of Hayao Miyazaki, featuring whimsical characters and vibrant landscapes \newline
D\&D, fantasy, elegant, pale, highly detailed, digital painting, artstation, concept art, illustration, art by alberto scorfano and james jean and jason chan \newline
painting by James Favaro \newline
Water Lilies, oil on canvas, Impressionist series capturing the serene beauty of Monet's garden pond \newline
painting by Vektroid, digital art \newline
Minimalist line art by Saul Steinberg, ink on paper \newline
Dark fantasy, Warhammer, ArtStation painted by Zdzisław Beksiński and Wayne Barlowe \newline
fauvism style painting, oil on canvas, vibrant colors \newline
Organic surrealism \newline
Cinematic, dark scenes with film grain and deep, moody color tones \newline
painting by Yves Klein, oil on canvas \newline
Pop art colorful graphics \newline
vaporwave aesthetic, synthwave, digital painting, artstation, concept art, smooth, sharp focus, art by artgerm and greg rutkowski and alphonse mucha \newline
painting by Jean-Honoré Fragonard \newline
Kitsch art by Jeff Koons \newline
Digital hyperrealism with intricate textures by contemporary artists using lifelike color schemes \newline
painting by Moebius \newline
painting by James Turrell \newline
Georgia O'Keeffe Modernist \newline
A fusion of Van Gogh, Picasso, Cezanne, and David Hockney styles, oil on canvas with expressive brushstrokes and bold geometry \newline
cinematic, highly detailed, masterpiece by craig mullins, ruan jia, greg rutkowski, jakub rebelka, caravaggio, syd mead \newline
Impressionist painting with soft pastel hues and loose, visible brushstrokes inspired by Claude Monet \newline
Digital hyperrealism with intricate textures by contemporary artists, digital painting with stunning detail \newline
Superflat acrylic artwork by Takashi Murakami, blending traditional Japanese motifs with contemporary pop culture \newline
generative art, sci-fi, highly detailed, 3d octane render, high contrast, minimalistic by paul lehr and ralph mcquarrie \newline
Ophelia by John Everett Millais \newline
Neo-Expressionist raw brushstrokes and emotive color palette conveying intense feelings \newline
Imagery that evokes surreal, dreamlike scenes \newline\\
\bottomrule
\end{tabular}
\end{table*}
\end{appendices}

\end{document}